\PassOptionsToPackage{super}{natbib}
\documentclass[pdflatex,sn-nature,oneside]{sn-jnl}

\usepackage{graphicx}
\usepackage{multirow}
\usepackage{amsmath,amssymb}
\usepackage{xcolor}
\usepackage{booktabs}
\usepackage{tabularx}
\usepackage[title]{appendix}
\usepackage{url}

\newcommand{\pro}{BioXArena-Pro}
\newcommand{\base}{BioXArena}

\usepackage{longtable,array,ragged2e}

\providecommand{\bibname}{References}
\begin{document}

\title[BioXArena-Pro: a research-grade benchmark for biomedical ML agents]{BioXArena-Pro: a research-grade benchmark for autonomous biomedical machine learning agents}

\author[1]{\fnm{Loka} \sur{Li}}
\author*[1]{\fnm{Duzhen} \sur{Zhang}}\email{duzhen.zhang@mbzuai.ac.ae}
\author[1]{\fnm{Xingbo} \sur{Du}}
\author[2]{\fnm{Leonard} \sur{Song}}
\author[1,3]{\fnm{Kun} \sur{Zhang}}
\author*[1]{\fnm{Le} \sur{Song}}\email{le.song@mbzuai.ac.ae}
\affil[1]{\orgname{Mohamed bin Zayed University of Artificial Intelligence}, \orgaddress{\city{Abu Dhabi}, \country{United Arab Emirates}}}
\affil[2]{\orgname{The Westminster Schools}, \orgaddress{\city{Atlanta}, \state{Georgia}, \country{United States}}}
\affil[3]{\orgname{Carnegie Mellon University}, \orgaddress{\city{Pittsburgh}, \state{Pennsylvania}, \country{United States}}}

\abstract{Coding agents are increasingly proposed as autonomous builders of biomedical models, but existing benchmarks are small, near saturation or open to leaked information. We present \pro{}, 64 research-level modelling tasks from eight biomedical domains, each rebuilt from a published study; most hold out whole biological units such as species or cohorts. Across 22 agent configurations, the best scores 55.1 of 100, and the scaffold matters almost as much as the model. We re-ran a published method for every task: where the published recipe applies unchanged, the best agent and the methods tie; on rebuilt tasks the methods trail by 8.2 points, suggesting that agents gain by adapting to new data. Pretrained models (including biological foundation models) matter where the strongest agent uses them: forbidding them decreases its score on those tasks by 9 points. Every task was audited in several AI-assisted rounds and a final human review.}


\maketitle

Large language model (LLM) agents can now inspect data, write and debug code, train models and iterate on
results with little human input\cite{yao2022react,wang2024codeact,jimenez2024swebench}, and in biomedicine they are being proposed as
research assistants and even autonomous scientists\cite{huang2025biomni,jin2025stella,gottweis2025coscientist}.
Whether this promise holds depends on measurement: not whether an agent produces a plausible-looking model, but whether
its held-out performance reflects real modelling ability on the data that biomedical researchers actually face.

Current evaluations fall short in two ways. First, most scientific and biomedical agent benchmarks test knowledge, tool
use or analysis traces rather than end-to-end model building\cite{laurent2024labbench,mitchener2025bixbench,chen2025scienceagentbench},
and the existing model-building benchmarks are small, drawn from competitions, or limited to clean tabular inputs and
random splits\cite{huang2024mlagentbench,chan2024mlebench,miller2025biomlbench}. NatureBench tests whether coding agents
can match the published state of the art of Nature-family papers across the sciences\cite{wang2026naturebench}; it
inspired our published-method references, which focus on biomedical machine learning. Our earlier benchmark, \base{}
(76 tasks in nine domains)\cite{bioxarena2026}, broadened the coverage, but it is approaching saturation: on the common
scale used here, the two strongest configurations already average 66 of 100, and on 12 of its 76 tasks the best of four
configurations comes within 10 points of the maximum. Second, a coding agent does more than fit a model. It reads the
task files, searches the data for regularities and tests what the grader rewards, so it will find and use any leaked
label, metadata artefact or grader loophole left in a benchmark, and the resulting score will overstate its
ability\cite{geirhos2020shortcut,kapoor2023leakage,skalse2022rewardhacking}. Biological data are especially prone to
such leaks, through homology or similarity between training and test sequences and structures, entities shared across splits and artefacts of data
generation\cite{whalen2022pitfalls,bernett2024ppi,buttenschoen2024posebusters}. A benchmark for these agents must
therefore be harder than current ones, and it must be checked for shortcuts before release.

Here we present \pro{} (Fig.~\ref{fig:overview}). Each of its 64 tasks reproduces a research-grade modelling problem
from a published study or database and keeps the distribution shift, data volume and native formats of the original
work; examples are predicting perturbation responses in unseen cell lines and transferring protein-interaction
predictors across species. Compared with \base{}, the median task carries about twice as much training data, and most
tasks hold out an entire biological unit, such as a species, cohort, donor or cell line, rather than random
rows\cite{koh2021wilds}. Every task is scored on a common 0--100 scale on which higher is better, and all capability
results come from the final release.

We report three results. First, we evaluated 22 configurations: Codex CLI\cite{openai2025codex}, Claude
Code\cite{anthropic2025claudecode} and Gemini CLI\cite{google2025geminicli}, each with two of its vendor's models, and
four model-agnostic scaffolds crossed with four models. The best configuration reaches 55.1 of 100 (54.5 on average over
three runs), no task is solved to saturation, and the scaffold around a model matters almost as much as the model
itself. Second, we re-ran a published method for every task under our protocol. For 33 tasks this is a method published
for the task's dataset; for the other 31 it is the strongest published method for the same problem class (Methods).
These references match the strongest agent where our test set stays close to the original study, and they trail it by
8.2 points on the tasks that we rebuilt against contamination. A published recipe is tuned to the data for which it was
designed, whereas the agents tailor validation, normalization and ensembling to each new split. Third, agents load
pretrained models, including biological foundation models\cite{lin2023esm2,cui2024scgpt}, in 18\% of their solutions.
Where the strongest agent uses them, they carry real signal: forbidding them decreases its score on those tasks by 9 points. All three results rest on
a benchmark that we audited in several rounds before release, because frontier agents had exploited integrity defects in earlier versions (Methods). We release the benchmark, the evaluation code, the leaderboard, the
audit protocol, the reproduced references and all agent traces.

\section{Results}\label{sec:results}

\subsection{A research-grade benchmark for biomedical modelling}\label{sec:benchmark}
\begin{figure}[!t]
\centering
\includegraphics[width=\textwidth]{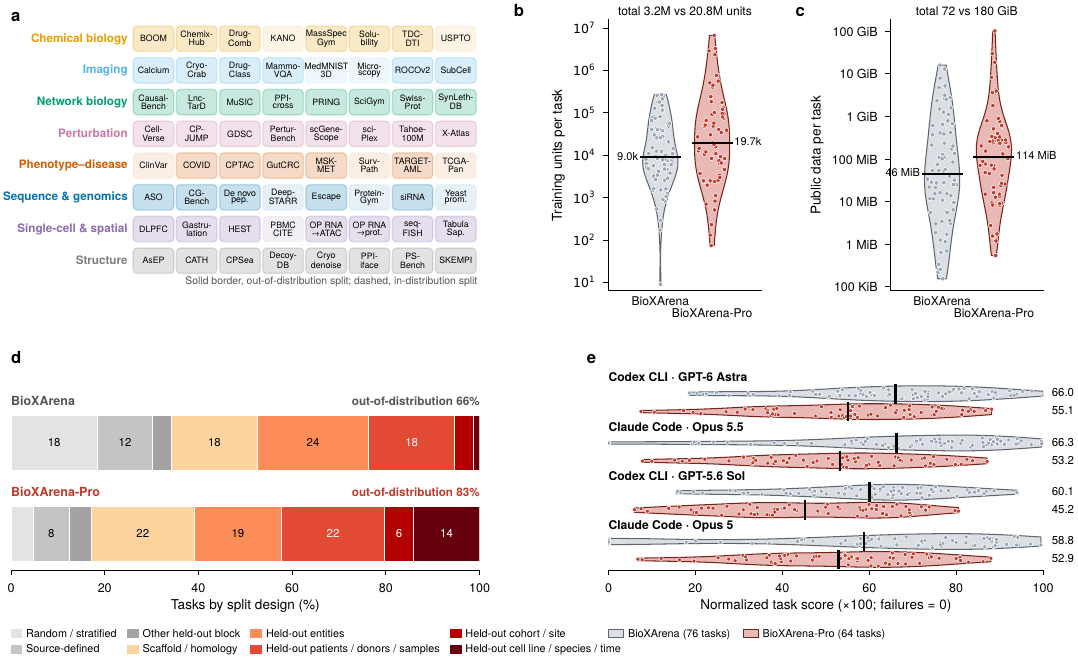}
\caption{\textbf{\pro{} is a larger, more heterogeneous and harder successor to \base{}.}
\textbf{a}, The 64 tasks, one row per domain; a solid border marks a test set that holds out a
biological or clinical unit (out-of-distribution split), a dashed border an in-distribution split.
\textbf{b}, Training units per task (log scale) for \base{} (grey, 76 tasks) and \pro{} (red, 64 tasks); violins show
the distribution, points are tasks, bars are medians and the title gives the total. \textbf{c}, Public data per task, drawn
as in \textbf{b}. \textbf{d}, Share of tasks by
split design, from random, source-defined or other in-distribution splits (greys) to held-out scaffolds, entities, donors, cohorts, cell lines,
species and time windows (reds); the label gives the out-of-distribution share. \textbf{e}, Normalized task scores ($\times 100$)
of the four agent configurations run on both benchmarks, on the common \pro{} scale (Methods); each point is one task,
bars are means and the number on the right is the mean over tasks.}
\label{fig:overview}
\end{figure}

\pro{} comprises 64 tasks, eight per domain, across chemical biology, imaging, network biology, perturbation biology,
phenotype--disease modelling, sequence and genomics, single-cell and spatial omics, and structural biology
(Fig.~\ref{fig:overview}a, Supplementary Table~1). Each task is a self-contained capsule: a public directory with
training data, unlabelled test inputs, native-format assets and a description; test labels are kept private. An agent receives the capsule, one GPU and a two-hour budget, and must return predictions for the test set
(Methods).

\begin{figure}[!t]
\centering
\includegraphics[width=\textwidth]{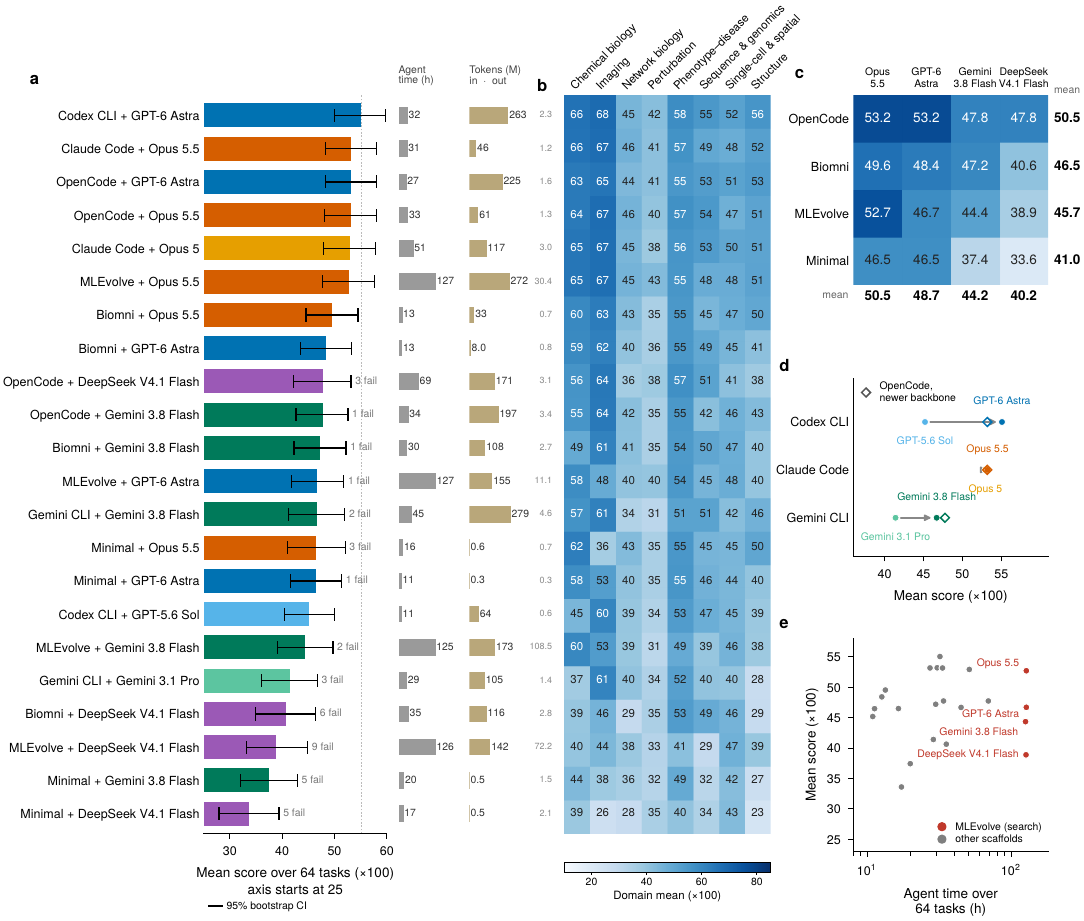}
\caption{\textbf{Capability of agent configurations on the final release.}
\textbf{a}, Mean score over the 64 tasks for each of the 22 configurations (bars; the axis starts at 25), with 95\%
bootstrap CIs over tasks (5,000 resamples); failed tasks score 0 and their count is given next to the interval. Dotted line, the best mean. The two columns to the right give the
agent wall-clock time summed over the 64 tasks and the input (bar) and output (small number) tokens in millions; input
includes cached prompt tokens. Colour marks the backbone (one colour family per vendor, the newer backbone darker).
\textbf{b}, Domain means per configuration, rows ordered as in \textbf{a} ($n = 8$ tasks per domain).
\textbf{c}, Scaffold--backbone design: mean score of four model-agnostic scaffolds with four backbones, with row and
column means. \textbf{d}, Vendor-CLI design: each coding CLI with the older and the newer of its vendor's two backbones
(arrow), and OpenCode with the newer backbone (open diamond). \textbf{e}, Agent wall-clock time against mean score;
red, the search-based MLEvolve scaffold.}
\label{fig:capability}
\end{figure}

Relative to \base{}, the tasks are larger, more heterogeneous and harder by construction
(Fig.~\ref{fig:overview}b--d). The median task provides 19,663 training units (\base{}: 9,042) and occupies 114\,MiB
(46\,MiB); the public data total 179.6\,GiB (71.6\,GiB). Only 16\% of tasks can be trained from a single train and test
table (\base{}: 50\%); the others link data files in native formats. A task draws on 3.25 input sources on average
(\base{}: 2.62 under the same definition), and 34\% of tasks draw on four or more (\base{}: 14\%). Over a third of tasks
(36\%) combine at least two families of input modality (45\% if borderline cases are counted; Supplementary Table~3).
Most tasks (83\%) use a split that holds out an entire group of related samples rather than random rows (\base{}: 66\%);
the held-out groups are mostly biological or clinical units, such as species, cohorts, donors, patients, cell lines and
protein families, and otherwise chemical scaffolds, property ranges or time windows. Metrics follow the scientific
output; 75\% of them are averaged per group, and every score is mapped to a common 0--100 scale on which higher is better
(Methods). For the four agent configurations that we ran on both benchmarks, mean scores on the same scale are 11 points
lower on \pro{} than on \base{} (51.6 versus 62.8; Fig.~\ref{fig:overview}e). The number of near-saturated tasks, with a
best score of at least 90, falls from 12 of 76 (best of the four configurations) to 0 of 64 (best of all 22).

\subsection{Capability of 22 agent configurations on the final release}\label{sec:capability}
On the final release we evaluated 22 agent configurations in two designs (Fig.~\ref{fig:capability}; Methods). The
first design pairs each of three vendor coding command-line interfaces (CLIs), namely Codex CLI, Claude Code and Gemini
CLI, with two backbones from its own vendor. The second design crosses four model-agnostic scaffolds with four backbones.
The scaffolds are an open-source coding CLI (OpenCode\cite{opencode2025}), a biomedical agent
(Biomni\cite{huang2025biomni}), a search-based agent for machine-learning engineering (MLEvolve\cite{du2026mlevolve}) and
a minimal scaffold that sends a single prompt; the backbones are the newer model of each vendor CLI and an open-weight
model, DeepSeek V4.1 Flash. Codex CLI with GPT-6 Astra reaches 55.1 (95\% confidence interval, CI, 50.0--59.8),
followed by Claude Code with Opus 5.5 and by OpenCode with GPT-6 Astra or Opus 5.5 (all 53.2); the minimal scaffold with
DeepSeek V4.1 Flash is last at 33.6 (Fig.~\ref{fig:capability}a). Domain matters about as much as configuration: domain
means range from 36.4 for perturbation biology to 56.4 for imaging (Fig.~\ref{fig:capability}b). Cost varies more than
score: agent time ranges from 11 to 127 hours and input tokens from 0.3 to 279 million, with little relation to rank
(Fig.~\ref{fig:capability}a,e; Supplementary Table~4).

In the four-by-four design, OpenCode is the best and the minimal scaffold the worst scaffold for every backbone
(Fig.~\ref{fig:capability}c); across the 16 cells, the backbone accounts for 53\% of the variance in mean score and the
scaffold for 37\%. A full coding scaffold adds more as the backbone weakens, from 6.7 points with GPT-6 Astra and
Opus 5.5 to 14.2 with DeepSeek V4.1 Flash. As a result, the open-weight backbone in OpenCode (47.8) is level with both
frontier backbones in the minimal scaffold (46.5; the difference is within the CIs). On their own backbone, vendor CLIs
hold little advantage over the open-source CLI (Fig.~\ref{fig:capability}d): Codex CLI exceeds OpenCode by 1.9 points
(95\% CI 1.1--2.7), and Claude Code and Gemini CLI differ from it by 0.0 and $-1.1$. Within a vendor CLI, the newer
backbone scores 9.9 points higher for GPT-6 Astra over GPT-5.6 Sol and 5.3 points higher for Gemini 3.8 Flash over
3.1 Pro (a comparison that also crosses model tiers), but only 0.3 points higher for Opus 5.5 over Opus 5.

More search did not help within the two-hour budget. MLEvolve iteratively mutates and re-evaluates candidate
solutions. With every backbone it spent about 127 agent-hours, four times as much as Codex CLI with GPT-6 Astra
(Fig.~\ref{fig:capability}e), yet it never beat the best coding CLI on the same backbone: it trailed by 8--9 points with
GPT-6 Astra and DeepSeek V4.1 Flash and did not differ significantly with Gemini 3.8 Flash ($-3.4$) or Opus 5.5
($-0.5$).

\begin{figure}[!t]
\centering
\includegraphics[width=\textwidth]{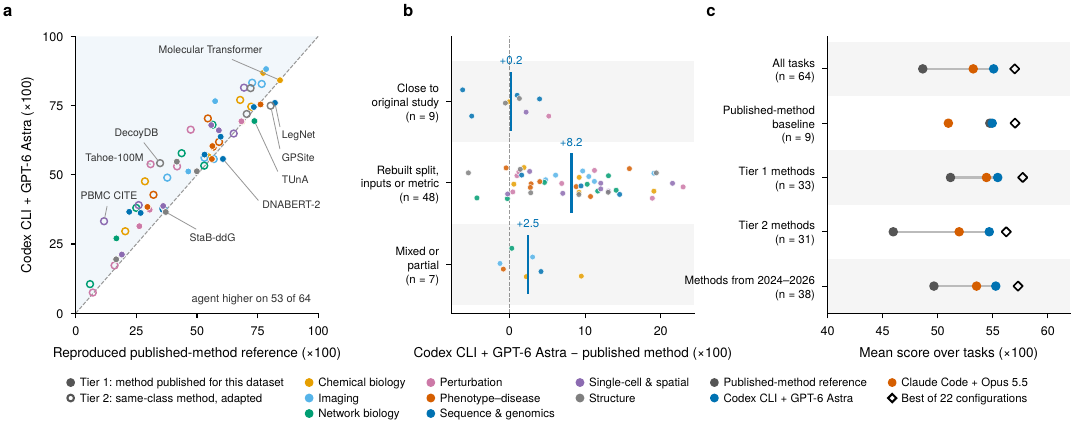}
\caption{\textbf{Reproduced published-method references match agents only where the test set stays close to the original
study.}
\textbf{a}, Per task, the reproduced published-method reference (the primary method, chosen by validation on training data)
against Codex CLI with GPT-6 Astra; colour marks the domain as in Fig.~\ref{fig:overview}, filled points are tier-1
methods (published for this dataset) and open points tier-2 methods (strongest published method for the same problem
class, adapted); labels name selected methods that beat the agent and tasks with the largest agent lead.
\textbf{b}, Agent minus published-method reference per task, grouped by how close our test set is to the original study: close (same data
and metric family), rebuilt (new split, removed inputs or new metric) or mixed (the grade differs between instances or applies to part of the task); bars are means.
\textbf{c}, Mean score over tasks for the published-method reference, the two strongest configurations and the per-task best of
the 22 configurations, for all tasks, the published-method baseline (tier-1 methods on tasks close to the original study) and the tier-1, tier-2 and 2024--2026 subsets. All scores $\times 100$.}
\label{fig:expert}
\end{figure}

Picking the best agent for each task would add little: the per-task best of all 22 configurations averages 57.0, only
1.9 points above the best single configuration, and the per-task winners are spread over 12 configurations. No
configuration comes within 10 points of the maximum on any task, and on three tasks the best of all 22 stays below 20.
With two further runs, the two strongest configurations average 54.5 $\pm$ 0.5 and 53.5 $\pm$ 0.3 (mean $\pm$ s.d. over three runs). The 64-task mean is therefore stable across runs, and the headline 55.1 is the highest of three, although
individual tasks can swing by more than 15 points.

\begin{figure}[!t]
\centering
\includegraphics[width=\textwidth]{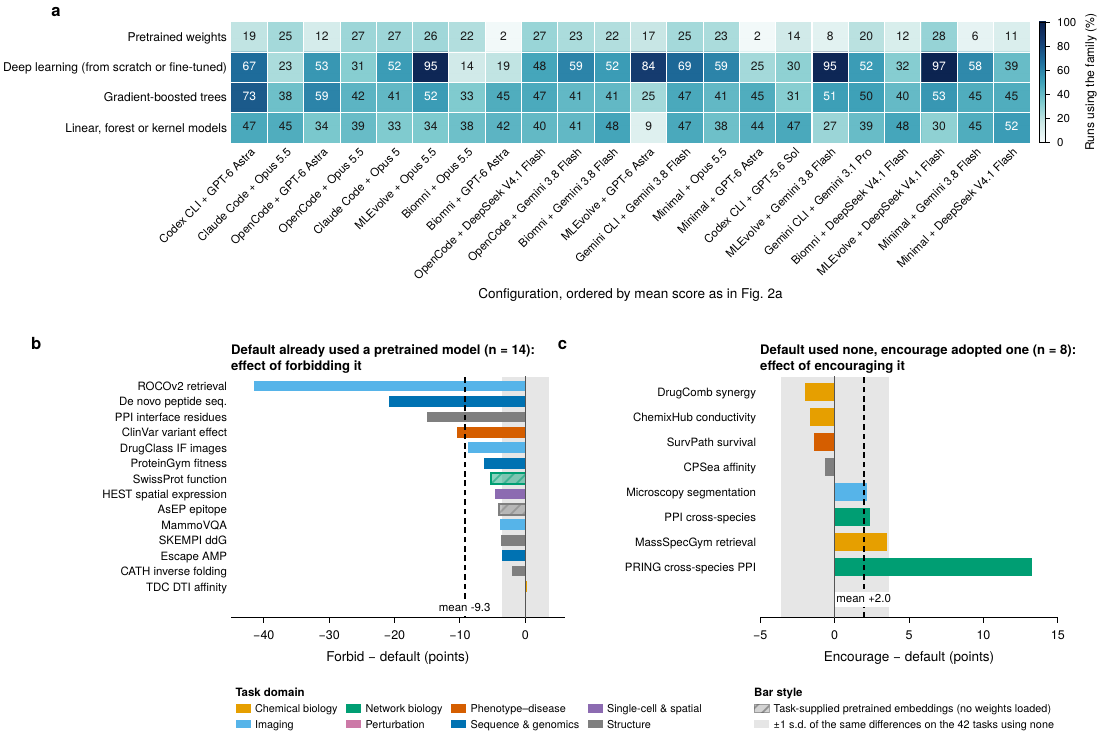}
\caption{\textbf{Pretrained models: what the configurations use and what they are worth.}
\textbf{a}, Share of runs whose final code uses each model family, per configuration (22 configurations $\times$ 64
tasks, 1,401 runs with code; families are not exclusive). \textbf{b}, Codex CLI with GPT-6 Astra on the 14 tasks whose
default solution already relied on a pretrained model: change in score when the prompt forbids all pretrained weights
and embeddings; hatched bars, tasks whose default used task-supplied pretrained embeddings rather than loaded weights.
\textbf{c}, The eight tasks whose default solution used no pretrained model and whose solution under an encouraging
prompt (with a local zoo of 99 models) adopted one: change in score. Dashed lines are means; the grey band is $\pm$1 s.d.
of the same differences on the 42 tasks with no pretrained model in either arm. Bars are coloured by domain as in
Fig.~\ref{fig:overview}.}
\label{fig:pretrained}
\end{figure}

\subsection{Agents against reproduced published-method references}\label{sec:expert}
To interpret absolute scores, we compared the agents with a reproduced published-method reference for every task
(Fig.~\ref{fig:expert}). For 33 tasks (tier 1) we re-ran a method published for the task's dataset. For the other 31
(tier 2) we re-ran the strongest published method for the same problem class, because no dataset-specific method
exists, none could be run legitimately or a same-class method validated better. Each method used the authors' code and
training recipe, was trained only on the public training data and used no checkpoint whose training data overlap the
test samples; the primary method was chosen by validation on training data and scored with the official grader
(Methods). These references (median publication year 2024) reach 48.6 on average, below Codex CLI with GPT-6 Astra
(55.1) and Claude Code with Opus 5.5 (53.2); the agent is higher on 53 of 64 tasks (Fig.~\ref{fig:expert}a). The gap is
concentrated in the tasks that we rebuilt against contamination (Fig.~\ref{fig:expert}b). Nine tasks combine a method
published for the dataset with a test set close to the original study, so the published recipe applies unchanged; we
treat them as the published-method baseline. On these nine tasks the published methods score 54.7 and the strongest
agent 54.9 (higher on five of nine; difference $+0.2$, 95\% bootstrap CI $-2.2$ to $2.4$), and Claude Code with
Opus 5.5 scores 51.0. On the 48 tasks rebuilt with new splits, removed inputs or new metrics, the references trail the
agent by 8.2 points (CI 6.2 to 10.1). The gap persists for recent and for dataset-specific methods: methods from
2024--2026 score 49.7 against 55.3 (38 tasks), and tier-1 methods 51.2 against 55.4 (Fig.~\ref{fig:expert}c). Published
methods still beat Codex CLI on 11 tasks and match or exceed the best of all 22 configurations on four, where
specialised architectures or pretraining are decisive (for example LegNet\cite{penzar2023legnet} on yeast promoter
activity and GPSite\cite{yuan2024gpsite} on protein-interface residues). The largest deficits, of 15 to 23 points, arise
where the rebuilt task removes an input on which the method depends.

\begin{figure}[!t]
\centering
\includegraphics[width=\textwidth]{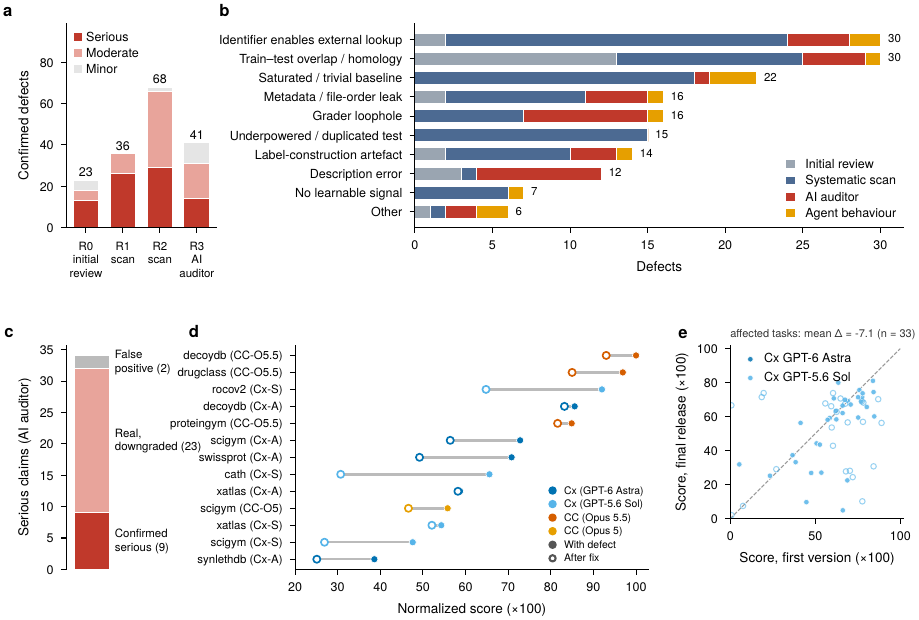}
\caption{\textbf{Integrity defects in pre-release builds and their exploitation by agents.}
\textbf{a}, Confirmed defects per audit round, by severity ($n = 168$ defects in 62 of 64 tasks); the 14 serious defects of round 3 are the auditor's nine confirmed claims of \textbf{c}, two further tasks that share a confirmed grader defect, one found by direct review of the task the auditor declined and two found while fixing. \textbf{b}, Defects by
category, coloured by who found them: the initial review, the systematic scans, the adversarial AI auditor or an anomalous
agent score or submission. \textbf{c}, Outcome of human verification for the 34 claims the AI auditor rated serious.
\textbf{d}, Agent scores with the defect (filled) and after the fix (open) for the 13 clear or partial exploitation cases with a comparable rerun (the score after a fix is that on the task version of that round, not on the final release); Cx, Codex CLI (A, GPT-6 Astra; S, GPT-5.6 Sol); CC, Claude Code (O5.5, Opus 5.5; O5, Opus 5). \textbf{e}, Score of the same configuration on the
first version and on the final release for every task with a valid run on both; filled points are tasks whose defects were
fixed without changing the metric or scale ($n = 33$ per configuration), open points the rest; the title gives the mean
change for Codex CLI with GPT-5.6 Sol ($-7.1$; $-3.9$ for Codex CLI with GPT-6 Astra on the same 33 tasks). These shifts combine the removal of defects with other hardening of the rebuilt tasks. Scores in \textbf{d} and \textbf{e} are $\times 100$.}
\label{fig:integrity}
\end{figure}

We read this as evidence that \pro{} rewards adapting a pipeline to each task's shift rather than reusing one. The agent transcripts contain per-batch normalization, validation splits that mimic the shift between training and test data,
ensembles and task-specific post-processing that no published recipe prescribes. This does not imply that agents
outperform human experts, who would adapt as well. The comparison also has asymmetries that favour neither side. The authors implemented the references from the published recipes, partly with the help of AI coding agents; half of the references are same-class methods; they used pretrained checkpoints only when these excluded our test samples; and they were not tuned to the rebuilt splits. The agents, in turn, had two hours and no domain recipe (Methods). Nor are domain methods dispensable: where they win, a domain-specific architecture or pretraining matters.

\subsection{Pretrained models help where agents already use them}\label{sec:pretrained}
We classified the final code of the 1,401 runs that left code (Fig.~\ref{fig:pretrained}a). Deep learning appears in
52\% of runs, gradient-boosted trees in 45\% and linear, forest or kernel models in 39\%, but only 18\% of runs load
pretrained weights, including biological foundation models. How often a configuration uses pretrained models depends on
both the backbone and the scaffold: Opus 5.5 loads them in 22--27\% of runs in every scaffold, whereas GPT-6 Astra does
so in 19\% of runs in Codex CLI but in 2\% in Biomni and in the minimal scaffold.

To measure what they are worth, we reran Codex CLI with GPT-6 Astra on all 64 tasks with two modified prompts: one
forbidding all pretrained weights and embeddings, and one pointing the agent to a local zoo of 99 pretrained models
(Methods). The two arms average 52.7 and 54.6 against 55.1 for the default prompt (54.5 on average over three runs), and the effect is asymmetric. On the
14 tasks whose default solution already relied on a pretrained model, forbidding it cost 9.3 points on average (lower on
13 of 14, median $-4.9$, Wilcoxon $P = 0.0002$), and up to 41.5 points for image--caption retrieval without BiomedCLIP\cite{zhang2025biomedclip}
(Fig.~\ref{fig:pretrained}b). On the eight tasks where the default solution used none and the encouraging prompt made
the agent adopt one, scores changed by $+2.0$ on average, four up and four down ($P = 0.31$); only cross-species
interaction prediction on the PRING-derived task gained clearly ($+13.3$ with ESM2 and ProtT5\cite{elnaggar2022prottrans}; Fig.~\ref{fig:pretrained}c). On the 42 tasks with no pretrained model in either arm the differences had an s.d. of
3.6 points, so single-task changes of a few points are noise. For this agent, pretrained models thus carry real signal on about a fifth of the tasks, and it already finds them there; telling it to use more does not help on average. The grouping is post hoc and each arm is a single run, so the per-task values are indicative.

\subsection{Frontier agents find and exploit integrity defects}\label{sec:exploit}
Before release we audited every task in four rounds, one manual and three AI-assisted, each followed by expert verification of the flagged
problems, and closed with a final expert review of every task and fix (Fig.~\ref{fig:integrity}a; Methods). The rounds
confirmed 168 defects in 62 of the 64 tasks (82 serious), most often identifiers or public files that let a networked
agent retrieve the answer, train--test overlap through homology or shared entities, and tasks that a trivial baseline
saturated (Fig.~\ref{fig:integrity}b). Agents exploited such defects without instruction: of 37 runs examined because of
an anomalous score or submission, 12 clearly exploited a defect and five in part (Fig.~\ref{fig:integrity}d shows the 13 of these 17 with a comparable rerun); of the AI auditor's 34 serious claims, verification confirmed nine (Fig.~\ref{fig:integrity}c). Claude Code
with Opus 5.5 reached a perfect 100 on a pose-classification task by learning a bond-length artefact that separated decoy
from native poses, and Codex CLI with GPT-6 Astra scored 72.8 on a network-completion task by reading gaps in element
numbering (56.4 after the fix). Across the 13 clear or partial cases with a comparable rerun, the fix lowered the same configuration's score by 13.1 points on average (range 0.5--34.9). These scores after a fix come from the task version of that round, which later rounds changed further. Across all tasks fixed without changing the metric, scores fell by 3.9--7.1 points (Fig.~\ref{fig:integrity}e), a shift that also includes other hardening of the rebuilt tasks. Because only runs with anomalous scores were examined, these counts do not estimate how often exploitation occurs; they show that it occurs without instruction and is hard to detect from aggregate scores. This is why all capability results use the final release.

\section{Discussion}\label{sec:discussion}
Coding agents can already build credible models for many biomedical problems: on the final release they produce valid
submissions in 97\% of the 1,408 runs (22 configurations on 64 tasks), and the strongest configurations exceed reproduced published
methods on most tasks. But the same competence makes them dangerous to evaluate naively: agents found and used label
leakage, metadata artefacts and grader loopholes that had passed expert curation.

We draw four lessons. First, a benchmark for data-reading agents needs opaque identifiers, controlled metadata,
group-aware splits, shortcut baselines and graders that cannot be gamed, and it needs an adversarial audit before
release; AI auditors are valuable red-teamers, but they over-report severity, so their claims need verification against
the real grading path. Second, what the benchmark measures is adaptation rather than reuse: published methods hold their
own where our test sets stay close to the original studies and fall behind where we introduced new splits, inputs or
metrics, which is also where biomedical models matter most, in new cohorts, species and cell lines. Third, a score
belongs to a scaffold--backbone pair: a weaker backbone in a strong scaffold can match a stronger backbone in a weak
one. Fourth, pretrained models matter on about a fifth of the tasks and add little elsewhere; the
strongest agent already uses them where they help, so the frontier is better judgement about when and how to use them,
not more encouragement.

The results also show where the frontier lies. The per-task best of all 22 configurations is only 1.9 points above the
best single configuration, so the remaining 43 points are not recoverable by picking among these agents. Perturbation biology and network biology stay below 41 on average, and three tasks remain under 20 for every configuration. These include problems that remain open in the field, such as predicting the effects of unseen genetic perturbations\cite{ahlmanneltze2025perturbation} or responses in unseen cell lines. Without noise ceilings, however, we cannot separate unsolved biology from label noise or from difficulty added by removing identifiers. More search within the same per-task limit does not close this gap either: the search-based scaffold spent four times the agent-hours of the best coding CLI for no gain. The lead of Codex CLI with GPT-6 Astra persists on three-run means (54.5 against 53.5), whereas the next five configurations are indistinguishable. We expect progress to come from agents that reason
about the data-generating process, choose or adapt domain models deliberately and verify their own validation design.

Our study has limitations. Lookup of public source data cannot be fully prevented when agents have network access: opaque identifiers and explicit prohibitions limit it, but inputs such as sequences or structures can themselves be searched, and network traffic was not captured. An audit of the logs of all 1,408 runs found no download of a source dataset and no database query with test inputs, but 40 runs used public checkpoints whose training data may overlap our test samples, and this overlap was not measured (Methods). The 0--100 scale is an index rather than a fraction solved. An uninformed submission scores zero on 36 tasks, 50 on the six AUC-type tasks and 8.5 on average over the 63 tasks where chance is defined (Supplementary Table~6), and no task has a measured noise ceiling; means across tasks and domains should therefore be read comparatively. De-identification removes prior knowledge that a domain method or a foundation model could legitimately use, which may lower both the references and the value of pretrained models. Backbones were served through different API routes. Most configurations were run once per task, and the differences among the top configurations are smaller than their bootstrap intervals. We did not measure human experts directly. The references are, however, expert-designed: each is a method that domain experts devised, tuned and published for that dataset or problem class, and where a published number on the same test set exists our reproductions match it (Methods). What a reference lacks is the adaptation an expert would make on meeting our rebuilt split, so it is best read as a lower bound on expert performance. The nine baseline tasks, where the published recipe needs no adaptation, give the most direct comparison: there the strongest agent and the published methods tie (54.9 against 54.7). The evidence therefore supports agents reaching the level of published methods on these problems, not exceeding what experts could achieve with the same freedom to adapt. The audits used models of the same families as the evaluated agents, and tasks were revised after agent runs had been inspected, so the final tasks may be harder for these families than for others. The prompt ablation covers one agent, and the 64 tasks sample eight domains.

\pro{} turns these lessons into a benchmark on which a score reflects modelling rather than lookup. As agents improve, new shortcuts will appear, so we release it as a versioned, living benchmark with a public audit log, an evaluation server that keeps test labels private and the tooling needed to audit new tasks.

\section{Methods}\label{sec:methods}

\subsection{Task selection and construction}
We first compiled a candidate list of datasets from recent primary studies in Nature, Science, Cell and their sister
journals and from the benchmarks of recent machine-learning venues, keeping those that release a dataset together with a
method and an evaluation. Three authors (L.L., D.Z. and X.D.) downloaded, processed and screened the candidates against five criteria. The
source had to be a primary study or curated database, preferably from the last few years (median source year 2024) and
with a peer-reviewed publication where one exists. Its licence had to permit academic non-commercial benchmark use. The
prediction target had to be scientifically meaningful, with a real downstream use. The task had to be difficult: we
preferred tasks that combine several input modalities or data sources and whose held-out unit (species, cohort, donor,
cell line, scaffold or time window) matches the scientific claim of the original study. Finally, the task had to fit one
GPU and two hours. Candidates that failed a criterion, duplicated a problem class already covered or saturated
under a trivial baseline were dropped or replaced during the integrity rounds described below. The resulting 64 tasks,
eight per domain, were reviewed at a high level by the two senior authors (K.Z. and Le Song) for relevance, balance and
challenge. \pro{} is not disjoint from \base{}. Of the 64 tasks, 45 share no source dataset with it, 8 reuse a \base{} dataset for a different prediction task, and 11 revisit a \base{} dataset and target with a new split, inputs and metric (six as whole tasks and five as one component of a multi-source task). In five of these 11 tasks, some test samples appear with labels, under other identifiers, in the public training data of \base{}: about 55\% of the test items of the sci-Plex task and of the Tox21 instance, and 7--14\% of those for Swiss-Prot function, two ClinVar instances and the ProteinGym stability assays. Each task keeps the native data formats of its source and exposes a fixed primary metric chosen to match the
scientific output. Per-task provenance, licence, split rule, modalities and published-method reference are listed in Supplementary
Tables~1--3.

\subsection{Data sources, licences and redistribution}
Every task is built from a public primary study or database, listed with its licence in
Supplementary Table~2 and, except for the ten tasks noted below, cited in its description. We followed the policy of \base{}: the packaging, splits, opaque identifiers, descriptions and
graders are released under CC BY-NC 4.0 for academic and other non-commercial use; sources must permit academic
non-commercial benchmark use; and sources whose terms restrict redistribution of the original records are released only
as processed, de-identified task files, with the original resource cited and left to be obtained from its provider.
Each task's data remain subject to the terms of its source. Three sources need a specific note. The MSK-MET task is
reproduced value for value from the authors' Zenodo deposit of the cohort (CC BY 4.0); because the same study is
distributed through cBioPortal, whose DataHub requires derived databases to be shared alike, the files of this task and
the cBioPortal-derived files of the paediatric leukaemia task are offered under ODbL 1.0. The leukaemia task uses only
open-access data of the TARGET programme, for which NCI requires no data-use certification; the TARGET purpose
limitations apply to controlled-access data, and use follows the NIH Genomic Data Sharing Policy. The ionic-liquid
conductivity values of one task derive from NIST Standard Reference Database 147 (ILThermo); they are shipped as
processed values for non-commercial research with the acknowledgement NIST requests, and permission for redistribution has
been requested from the NIST Standard Reference Data Program. Ten task descriptions deliberately withhold the name of
their source to limit answer lookup; their sources are disclosed in Supplementary Table~2.

\subsection{Ethics and data governance}
This study analysed only previously published, de-identified data and involved no human participants: we recruited no
participants, generated no new human or animal data and had no contact with any data subject. The 27 tasks that include human-derived data
(clinical cohorts, tumour profiles, medical images, tissue sections and cell atlases) use only resources that their
providers distribute openly in de-identified form, collected under the ethical approvals and consent procedures reported
in the original studies (Supplementary Table~2); no controlled-access data were used. We made no attempt to re-identify
individuals or to link records across sources. The released task files go further than the sources: sample, patient and
donor identifiers are replaced by opaque codes, free-text and date fields are removed, and clinical variables are kept
only where the task needs them and in the coarsened form in which the source released them. Users of the benchmark must comply with the terms of each source, and we will remove or replace any task on request of a data provider. Two tasks on oligonucleotide knockdown efficacy triggered a model's biosafety
refusal (see Agent configurations and execution environment); no task involves pathogen enhancement, toxin design or other dual-use content.

\subsection{Task capsules, submission contract and scoring}
Every task is a self-contained capsule with a public and a private part. The public directory holds
\texttt{description.md}, \texttt{train.csv}, \texttt{test.csv}, \texttt{sample\_submission.csv} and the task's
native-format assets (images, volumes, structures, expression matrices, graphs or embeddings). For 43 tasks the
prediction is a value per test row; the other 21 use a file manifest, in which each row of the submission names a
prediction file in the agent's output directory so that predictions keep their natural granularity (for example a mask
per image or an expression matrix per tissue section). The private directory holds \texttt{answers.csv}, any private
label files and the grader. A submission must reproduce the sample submission's rows, columns and identifier order; the submission is invalid, and the run a failure rather than a valid submission with a low score, if a file is missing or malformed, the schema or identifier set is wrong, a path is unsafe or a required value is not finite. With a public pre-flight check, an agent can catch packaging errors before the deadline. Each grader computes one fixed primary metric; task instances that
share a grader are averaged with equal weight unless the task description states other weights; the metric, its aggregation and the measured chance level of every task are given in Supplementary Table~6. Scores are normalized to a common 0--100 scale on which higher is better. Correlation coefficients and Matthews correlation are clipped to $[0, 1]$, so that chance scores zero, as it does for the skill metrics. AUC, C-index, average precision, $F_1$, accuracy and mean reciprocal rank keep their native $[0, 1]$ scale. Two perturbation tasks are expressed relative to the all-zero prediction, and the hospitalization forecast uses a ranked probability skill score. All scores are multiplied by 100. A failed run scores 0. For Fig.~\ref{fig:overview}e the 28 correlation-scored tasks of
\base{} were rescored from the raw metric with the same clip so that both benchmarks share one scale.

\subsection{Integrity hardening protocol}
Every task was audited in four rounds before release (Fig.~\ref{fig:integrity}a). AI agents did the searching in three of them; an author verified every AI finding and approved every fix. Round 0 was a manual review of the first version by the authors who constructed the tasks. Round 1 was a systematic scan with
Claude Code and Opus 5.5 for identifier leakage, train--test overlap and metadata--label correlations. Round 2 was a
second, broader systematic scan with Claude Code and Fable 5.1, extended to lookup of identifiers in public databases,
file-order and file-metadata leakage, homology and scaffold overlap, target-derived features and grader loopholes. Round 3
was an adversarial audit by a different agent, Codex CLI with GPT-6 Astra, which attacked the real grader of each task. In addition, anomalous agent scores and submissions from the evaluation runs were traced back to their
cause. After the last round, the authors manually re-examined all 64 tasks --- data files, splits, descriptions and
graders --- and signed off each one before the final release.

The rounds confirmed 168 defects in 62 of the 64 tasks (82 serious, 69 moderate, 17 minor), counted once per mechanism
per task. The largest classes were identifiers or public files that let a networked agent retrieve the answer (30),
train--test overlap through homology or shared entities (30), tasks that a trivial baseline saturated (22), metadata or
target-derived features correlated with labels (16) and grader loopholes (16). Most defects were fixed by rebuilding the
task, the others by changing the grader or rewriting the description; four tasks were retired and replaced, and one instance was withdrawn from another task. The 168 defects include eight in the four retired tasks; the 62 of 64 refers to the 64 tasks of the first version, and a defect found through agent behaviour is assigned to the round in which it was found. The fixes follow fixed rules. Identifiers are replaced by opaque salted codes, and lookup keys (gene symbols, variant strings, compound names and structures, accession numbers) are removed from test items. File order is shuffled and file metadata made uniform. Test sets are re-split so that an entire group of related samples is held out, with bounded sequence identity or disjoint scaffolds between training and test sets where applicable (Supplementary Table~1). Graders require exact coverage of the test identifiers and reject non-finite values and abstention, and descriptions state which external data and lookups are not allowed. The 37 examined runs were selected because a score or a submission was anomalous;
each was classified from its transcript and code, and the comparison in Fig.~\ref{fig:integrity}d uses the same
configuration rerun on the fixed task.

The round-3 auditor ran once per task in a sandbox with read-only access to the task's public and private directories
and its grader. It was asked to find ways of obtaining the labels or of scoring well without learning the biology, and
to measure each against the real grader. Of its 34 serious claims, re-execution by an author through the real grading
path confirmed nine, downgraded 23 whose gain was within noise and rejected two (Fig.~\ref{fig:integrity}c). Because
defects are counted per affected task, round 3 contributes 14 serious defects to Fig.~\ref{fig:integrity}a: the nine
confirmed claims, two further tasks that share a confirmed grader defect, one defect found by direct review of the task
that the auditor's safety filter declined, and two found by the authors while fixing.

\subsection{Agent configurations and execution environment}
We report 22 configurations in two designs. In the vendor-CLI design, each coding CLI runs with two backbones from its own
vendor: Codex CLI 0.163.0 with GPT-6 Astra and GPT-5.6 Sol, Claude Code 2.1.302 with Opus 5.5 and Opus 5, and Gemini CLI
with Gemini 3.8 Flash and Gemini 3.1 Pro (preview). In the scaffold--backbone design, four model-agnostic
scaffolds --- OpenCode 1.18.5, Biomni, MLEvolve and a minimal scaffold --- each run with GPT-6 Astra, Opus 5.5, Gemini 3.8
Flash and DeepSeek V4.1 Flash, that is, with the newer backbone of each vendor CLI and one open-weight model. The
minimal scaffold sends the task once and executes the single code block it returns, with no tools and no feedback, so it
measures what a backbone achieves without an agent loop. Three further configurations on the public leaderboard (Codex
CLI with GPT-6.1 Sol and the minimal scaffold with DeepSeek V4 Pro and with Gemini 3.1 Pro) fit neither design and are not
analysed here.

All configurations receive the same task prompt, one NVIDIA A100 GPU (80\,GB), 16 cores of an AMD EPYC 7713P processor, 96\,GB RAM and a 7,200\,s wall-clock limit per task. Each run is one Slurm job inside a filesystem sandbox, in which the task's public directory is read-only and only the run's own output directory is writable; labels and answer files are never on the agent's filesystem. Network access is not blocked, because
agents must reach the model API and may install packages and download pretrained weights; leakage of answers is instead
prevented by three safeguards. First, the Landlock filesystem sandbox restricts an agent to reading its task's public directory and the runtime libraries and to writing its own output directory, so private answers, graders, other tasks
and earlier runs are outside its view. Second, the data are de-identified as described under Integrity hardening protocol, so that test items cannot be matched to their public source records; some tasks also rotate spatial coordinates or project embeddings at random. Third, the shared task prompt forbids access to private
data, labels, answer files, graders and any non-public artefact and states that original data releases are out of
bounds, and 62 of the 64 task descriptions also state which external data and lookups are not allowed for that
task. A run that exceeds
the time limit, returns a missing or invalid submission or is killed for exceeding memory scores 0. Infrastructure
failures --- gateway errors, empty completions from the serving API, network drops and runs killed by another job --- are
rerun, and the replaced attempts are archived with the release. Every API response is logged together with the model that
served it, and a run is rejected if fewer than 95\% of its responses come from the requested model. The main evaluation
of the 22 configurations used 1,023 hours of agent wall-clock time, each run with one GPU allocated, and 2.54 billion input and 0.26 billion output tokens;
the published-method references, the prompt ablation and the repeat runs are not included in these totals.

GPT and Claude backbones were served through an institutional API proxy, and DeepSeek V4.1 Flash and Biomni with
Gemini 3.8 Flash through OpenRouter. The other five Gemini configurations ran with the same harness, prompt, tasks, limits
and hardware class through a gateway for the native Gemini protocol; ten of their runs failed on gateway errors or
external process kills and were rerun, two of them (OpenCode) through OpenRouter. For MLEvolve with Opus 5.5, the proxy
returned empty completions for two tasks (siRNA and antisense-oligonucleotide knockdown efficacy), so these two scores
are taken from MLEvolve with Opus 5; without them the configuration's mean is 51.3 instead of 52.7. On the same two
tasks, Claude Code with Opus 5.5 declined on biosafety grounds, and the CLI's built-in refusal fallback completed them
with Opus 5; these are the only runs exempted from the 95\% rule.

\subsection{Audit of network use}
Because agents had network access, we audited all 1,408 runs of the final release for network use, from the recorded transcripts, the executed code, the scripts that the agents wrote and the per-run download caches. Pattern matching produced candidate events, and every candidate that involved a data repository, a database, a web search or a code repository was read manually. In total 362 runs (26\%) used the network: 84 installed packages and 313 requested pretrained weights (35 did both). No run downloaded the source dataset of its task, accessed any public data repository (including that of \base{}) or queried a public database with test inputs. The 313 runs include downloads that the final solution did not use, which is why the number exceeds the 18\% of final solutions that load pretrained weights. Seven runs searched for the source study by name; six retrieved nothing, and one read the code repository and paper of the source benchmark without retrieving data and scored below the median of the other configurations on that task. Forty runs used a public checkpoint whose training data may overlap the test samples: inverse-folding models on the CATH test chains, generalist Cellpose models\cite{stringer2021cellpose} on one of two test image sources, pathology encoders pretrained on TCGA images (inputs only, no survival labels) and one peptide-sequencing checkpoint of unverified provenance. The task descriptions permit these model classes, the overlap was not measured, and the runs are reported without exclusion. The audit covers what the logs record; network traffic itself was not captured.

\subsection{Reproduced published-method references}
For each task we reproduced a published method. Tier 1 is a method published for the task's dataset (33 tasks); tier 2
is the strongest published method for the same problem class, with only the input and output layers adapted (31 tasks).
A task is tier 2 when no dataset-specific method exists (11 tasks), when none could be run legitimately, or when a
same-class method validated better. Methods were run with the authors' code, version and training recipe and were
trained only on the task's public training data. Pretrained weights were allowed only when their training data contain
none of our test samples; otherwise the model was retrained from our data. Excluded checkpoints include
GNINA\cite{mcnutt2021gnina}, BepiPred-3.0\cite{clifford2022bepipred3}, the released PLM-interact
models\cite{plminteract2025}, STNeuroNet\cite{soltanianzadeh2019stneuronet}, LocalTransform\cite{chen2022localtransform}
and the Tahoe-specific perturbation models. Methods without available training code, or whose released model had been
trained on a resource that contains our labels, were not used; the code repository lists them per task. The authors implemented the references, partly with the help of AI coding agents, and every method directory records the paper, code commit,
weights and their provenance, hyperparameters, deviations, validation score and runtime. One to six methods were run per
task (median two). The primary method was chosen solely by validation on a split of the training data that mimics the
shift between training and test data, and test labels were never used to select or tune a method. For the
mutation-effect instance of the cross-species interaction task, fine-tuning was shortened from 50 to 23 epochs for
compute reasons, with the epoch selected by development-set AUPR as in the original recipe.

All 167 submissions were scored with the official grader. After a first scoring pass, we re-examined the nine tasks with the lowest reference scores for implementation errors; this is the only use made of test scores. We found and fixed
configuration errors in five of them (a merge step that never ran, a tree depth tuned for a random split, a splice model
chosen for the wrong organism, an unsuited classifier and an undertrained model) and confirmed the other four as
genuine; the primary method was then re-selected by validation. We report the primary method; the best method per task
by test score averages 49.0. Comparability with the original study was graded per task as close (same data and metric
family; 9 tasks), rebuilt (new split, removed inputs or new metric; 48 tasks) or mixed (the grade differs between
instances or applies to part of the task; 7 tasks). All nine close tasks have tier-1 methods. We call them the
published-method baseline, because the published recipe applies to them without adaptation (Supplementary Table~7), and
we treat the other 55 references as lower bounds on expert performance. Publication years come from the method papers.
Reproduction fidelity was checked for three methods with a published number on the same or nearly the same test set:
LegNet (the primary method) reaches 82.1 on the yeast promoter task, whose test set is drawn from that of the DREAM 2022
challenge, against about 82 reported; DeepGO-SE\cite{kulmanov2024deepgose} reaches 51.9 on Swiss-Prot function
prediction against 52.9; and Suite2p\cite{pachitariu2017suite2p} reaches an F1 of 0.504 on the real calcium-imaging
instance against the organisers' 0.499.

\subsection{Pretrained-model use and prompt ablation}
Model families were detected by regular expressions over the Python files that each run left in its output directory, excluding vendored packages: deep-learning frameworks, gradient-boosted tree libraries, linear, forest and kernel estimators, and the loading of pretrained weights (library loaders, hub downloads and names of common biological and vision models). Seven runs left no code and are excluded. The prompt ablation reran Codex CLI with GPT-6 Astra on all 64 tasks of the final release with the same harness, budget and hardware as the main evaluation. It used two modified prompts: the forbid prompt bans all pretrained weights and pretrained
embeddings, including embeddings shipped with a task; the encourage prompt adds a paragraph recommending pretrained
models and gives the path of a local zoo of 99 models, while stating that each task's own rules take precedence. The
default arm is the main run. Whether each solution loaded pretrained weights was decided by reading its code and
transcript; the automatic detector missed two task-specific pretrained models (InstaNovo\cite{instanovo2025} and ProteinMPNN\cite{dauparas2022proteinmpnn} with ESM2),
which were corrected by hand. Compliance of the forbid arm was checked for all 64 runs; none loaded, downloaded or
referenced pretrained weights, and the task-supplied ESM2 and IgFold embeddings were explicitly discarded.

\subsection{Statistical analysis}
Confidence intervals are 95\% percentile intervals from 5,000 bootstrap resamples of the 64 tasks; differences between
two configurations are bootstrapped paired by task. In the scaffold--backbone design, the shares of variance attributed to
scaffold, backbone and interaction come from a two-way decomposition of the 16 configuration means (one value per cell).
The per-task best is the mean over tasks of the highest score that any configuration reached on each task. To measure run-to-run variation, we ran Codex CLI with GPT-6 Astra and Claude Code with Opus 5.5 twice more on all 64 tasks, with the same prompt, pinned CLI versions, limits and API route as in the main round. We report the mean and s.d. of the three 64-task
means and the median over tasks of the per-task s.d. (1.3 and 1.1 points). In both repeat runs of Claude Code, the refusal fallback completed the same two tasks as in the main round. We tested 15 comparisons between pairs of configurations, which follow from the two designs and were fixed before the repeat runs and the reference reproductions: each vendor CLI against OpenCode on the same backbone, OpenCode against the minimal scaffold for each backbone, the newer against the older backbone in each vendor CLI, MLEvolve against the best coding CLI on each backbone, and the two strongest configurations against each other. For each comparison we report the mean paired difference, the number of tasks won and lost and a two-sided Wilcoxon signed-rank test
over the 64 tasks, with Holm correction across the 15 comparisons (Supplementary Table~5). All differences reported as such in Results remain significant after correction ($P_{\mathrm{Holm}} < 0.05$); the five comparisons described as small or not significant do not reach it. In the prompt ablation, the per-task score changes of each task group were tested against zero with a two-sided Wilcoxon signed-rank test. No other hypothesis tests were performed.

\subsection{Use of large language models}
Besides being the systems under evaluation, LLM-based coding agents assisted the authors in four ways, always under
author supervision. They carried out the searching in the three AI-assisted audit rounds (Claude Code with Opus 5.5 and
Fable 5.1, and Codex CLI with GPT-6 Astra; see Integrity hardening protocol). They partially helped implement the published-method
references from the authors' code and recipes. Claude Code with Opus 5.5 and Fable 5.2 assisted with analysis code, figures and reference checking. The same tools were used to edit the manuscript text. The authors
checked every output, verified each reported number against the underlying data and take full responsibility for the
content of this article.

\backmatter
\bmhead{Data availability} The public task capsules of \pro{} v5.1.2 (training data, unlabelled test inputs, native-format
assets, task descriptions and sample submissions for all 64 tasks; 179.6\,GiB) are available on Hugging Face at
\url{https://huggingface.co/datasets/mbzuai-ai4bio/BioXArena-Pro-Public}, as one archive and as one archive per domain,
under CC BY-NC 4.0 for academic and other non-commercial use; the source and licence of every task are listed in Supplementary Table~2 (ten descriptions withhold the source name to limit lookup), and each source's own licence also applies. The held-out test labels and answer files are not public, so that the benchmark cannot be contaminated; scores on the
test sets are obtained by submitting predictions to the \pro{} competition on Codabench\cite{xu2022codabench} (\url{https://www.codabench.org/competitions/18583/}), as for \base{}; during peer review the editor can give referees access to the test labels or to a scoring endpoint, and afterwards the labels are available from the corresponding authors to maintainers of evaluation infrastructure. Per-task scores of all 22 configurations are in the code repository, and the agent run records (the code written by each agent, its transcript and its submission) are available from the corresponding authors on request. The published-method references and the prompts and runs of the prompt ablation will be deposited with the versioned archive on acceptance. The detailed integrity-audit records, which list the shortcuts removed from each task, are available to referees and to maintainers of evaluation infrastructure on request, because publishing them would help future agents to exploit residual weaknesses.
\bmhead{Code availability} The benchmark harness, the adapters for the seven agent scaffolds (Codex CLI, Claude Code,
Gemini CLI, OpenCode, Biomni, MLEvolve and the minimal scaffold), the shared task prompt, the graders and normalization code, the scoring and leaderboard scripts, and the leaderboard files are available at \url{https://github.com/mbzuai-ai4bio/BioXArena-Pro} under the MIT License; a versioned archive will be
deposited at Zenodo on acceptance, together with the figure scripts and the directories of the published-method references.
\bmhead{Author contributions} L.L., D.Z. and X.D. collected, downloaded, processed and screened the datasets and built
the tasks. L.L. and D.Z. ran the experiments. L.L. wrote the first draft of the manuscript. D.Z., X.D., Leonard Song, K.Z. and Le Song revised the
manuscript and contributed to the analyses. K.Z. and Le Song supervised the project as senior advisors and guided its design and direction.
All authors read and approved the manuscript.
\bmhead{Competing interests} The authors declare no competing interests.

\clearpage
\newgeometry{margin=18mm,top=22mm,bottom=22mm}
\setcounter{table}{0}\setcounter{page}{1}\renewcommand{\thepage}{S\arabic{page}}
\renewcommand{\tablename}{Supplementary Table}
\makeatletter
\long\def\@makecaption#1#2{\vskip4pt{\footnotesize\raggedright{\bfseries #1\ $|$\ }#2\par}\vskip4pt}
\def\LT@makecaption#1#2#3{\LT@mcol\LT@cols{@{}l@{}}{\parbox[t]{\linewidth}{\footnotesize\raggedright #1{\bfseries #2\ $|$\ }#3\par\vskip 5pt}}}
\makeatother

\thispagestyle{empty}
\begin{center}
{\LARGE\bfseries Supplementary Information}\\[14pt]
{\Large \pro{}: a research-grade benchmark for autonomous biomedical machine learning agents}\\[14pt]
{\large Loka Li, Duzhen Zhang, Xingbo Du, Leonard Song, Kun Zhang and Le Song}
\end{center}
\vspace{18pt}
\noindent\textbf{Contents}\par\smallskip
\begin{tabular}{@{}p{0.86\linewidth}r@{}}
Supplementary Table 1 \ $|$ \ Task catalogue, data statistics and splits & \pageref{tab:catalog}\\
Supplementary Table 2 \ $|$ \ Data sources, references and licences & \pageref{tab:sources}\\
Supplementary Table 3 \ $|$ \ Input sources and modality families & \pageref{tab:sources_modalities}\\
Supplementary Table 4 \ $|$ \ The 22 agent configurations & \pageref{tab:configs}\\
Supplementary Table 5 \ $|$ \ Paired comparisons between configurations & \pageref{tab:paired}\\
Supplementary Table 6 \ $|$ \ Primary metric, normalization and chance level of each task & \pageref{tab:metrics}\\
Supplementary Table 7 \ $|$ \ Per-task scores of published-method references and agents & \pageref{tab:pertask}\\
Supplementary References & \pageref{LastPage}\\
\end{tabular}
\clearpage
{\scriptsize
\setlength{\tabcolsep}{2pt}
\setlength{\emergencystretch}{1em}
\begin{longtable}{@{}>{\RaggedRight\arraybackslash}p{0.125\linewidth}
>{\RaggedRight\arraybackslash}p{0.08\linewidth}
>{\RaggedRight\arraybackslash}p{0.155\linewidth}
>{\RaggedRight\arraybackslash}p{0.09\linewidth}
>{\RaggedRight\arraybackslash}p{0.15\linewidth}
>{\RaggedRight\arraybackslash}p{0.055\linewidth}
>{\RaggedRight\arraybackslash}p{0.14\linewidth}
>{\RaggedRight\arraybackslash}p{0.125\linewidth}@{}}
\caption{\textbf{Complete task catalogue of \pro{} with data statistics, split construction and reproduced published-method references.}}
\label{tab:catalog} \\
\multicolumn{8}{@{}p{0.97\linewidth}@{}}{``Type'' is the format of the predicted output, not the metric (Supplementary Table~6 gives the metric); abbreviations: Regr., regression; Segm., segmentation; Gener., generation. ``Train/Test'' gives the packaged prediction units (rows of \texttt{train.csv}/\texttt{test.csv}, or instances counted from the data files for index-style tables; `--' where the training set has no separate instance count). ``Split'' records the held-out unit as stated in the task description; $\circ$ marks an in-distribution split. ``Public size'' is the on-disk size of the task's public directory. ``Published method'' is the primary reproduced published method (tier 1, published for this dataset; tier 2, strongest published method for the problem class) and its score on the task ($\times 100$). Sources are given as in the task descriptions; full citations and licences are listed in Supplementary Table~2.} \\[2pt]
\toprule Task & Type & Metric & Train/Test & Split & Size & Source & Published method (tier, score) \\ \midrule
\endfirsthead
\caption[]{\textbf{Complete task catalogue of \pro{} (continued).}} \\
\toprule Task & Type & Metric & Train/Test & Split & Size & Source & Published method (tier, score) \\ \midrule
\endhead
\midrule \multicolumn{8}{r}{\textit{continued on next page}} \\
\endfoot
\bottomrule
\endlastfoot
\multicolumn{8}{@{}l}{\textbf{Chemical biology (8 tasks)}} \\
boom-csd & Regr. & mean of clip(r)+calibrated RMSE per source x tail & 15{,}925\,/\allowbreak\,2{,}000 & property-tail OOD (extrapolation) & 1.52M & BOOM (Antoniuk et al. 2025) density/HOF & boom\_\allowbreak{}egnn (1, 77.3) \\
chemixhub-conductivity & Regr. & residual clip(r)+calibration vs reference per source & 39{,}089\,/\allowbreak\,4{,}958 & group-disjoint (component set) & 8.13M & CheMixHub 2025; ionic-liquid DB; polymer-electrolyte compilation 2023 & chemeleon\_\allowbreak{}xgboost (1, 54.8) \\
drugcomb-multiomics-synergy & Regr. & projected Pearson per tissue mean & 176{,}729\,/\allowbreak\,93{,}964 & cold-drug & 243M & DrugComb v1.4 2021; NCI-ALMANAC 2017; CellMiner & deepsynergy (2, 28.5) \\
kano-molecular-property-prediction & Multi-\allowbreak{}label & AUPRC gain / R2 per target per instance & 183{,}881\,/\allowbreak\,22{,}996 & unspecified (KANO splits) $\circ$ & 245M & MoleculeNet via KANO (Fang 2023); PubChem subset; ElementKG & kano (1, 68.6) \\
massspecgym-msms-retrieval & Retrieval & MRR & --\,/\allowbreak\,628{,}956 & group-disjoint (MCES structural distance) & 280M & MassSpecGym 2024 (v1.5 2026) & metfrag (2, 67.8) \\
solubility & Regr. & clip(r)+calibrated RMSE & 6{,}452\,/\allowbreak\,3{,}527 & scaffold & 556K & AqSolDB (Sorkun 2019); Delaney 2004 & chemprop\_\allowbreak{}dmpnn (2, 72.4) \\
tdc-dti-binding-affinity & Regr. & weighted per-source clip(r)+calibration & 91{,}525\,/\allowbreak\,10{,}475 & cold-drug + cold-target (double-cold similarity purge) & 83.4M & TDC 2021: KIBA; DAVIS; BindingDB & psichic (2, 20.4) \\
\end{longtable}}

\clearpage
{\scriptsize
\setlength{\tabcolsep}{2pt}
\setlength{\emergencystretch}{1em}
\renewcommand{\UrlBreaks}{\do\/\do\-\do\.\do\_\do\=\do\?\do\&}
\begin{longtable}{@{}>{\RaggedRight\arraybackslash}p{0.13\linewidth}
>{\RaggedRight\arraybackslash}p{0.28\linewidth}
>{\RaggedRight\arraybackslash}p{0.075\linewidth}
>{\RaggedRight\arraybackslash}p{0.23\linewidth}
>{\RaggedRight\arraybackslash}p{0.245\linewidth}@{}}
\caption{\textbf{Primary data sources, references and licences of the 64 \pro{} tasks.}}
\label{tab:sources} \\
\multicolumn{5}{@{}p{0.97\linewidth}@{}}{One line per source; tasks built from several sources list them in order. ``Licence'' is the licence or terms of use stated on the primary release page of each source (data portal, Zenodo, Figshare, Dataverse, Hugging Face dataset card or repository LICENSE file), checked on 2026-11-09; where only a software licence is stated we say so. $^{\dagger}$ marks a licence we could not verify (no licence stated by the source, or the primary page could not be read); $^{\ddagger}$ marks a conflict between the task description shipped with the data and the primary release (wrong citation, different licence, or a record that belongs to another study), with the primary release given here. ``NC'' flags non-commercial terms. Full citation strings, the licence evidence and the redistribution notes are in \texttt{task\_sources.csv} in the code repository.} \\[2pt]
\toprule Task & Source dataset(s) & Ref. & Licence & Access \\ \midrule
\endfirsthead
\caption[]{\textbf{Primary data sources, references and licences of the 64 \pro{} tasks (continued).}} \\
\toprule Task & Source dataset(s) & Ref. & Licence & Access \\ \midrule
\endhead
\midrule \multicolumn{5}{r}{\textit{continued on next page}} \\
\endfoot
\bottomrule
\endlastfoot
\multicolumn{5}{@{}l}{\textbf{Chemical biology (8 tasks)}} \\
boom-\allowbreak{}csd & BOOM (CSD density / heat of formation) & \cite{S:boom2025} & Apache-2.0 (repository; no separate data licence)$^{\ddagger}$ & \url{https://github.com/FLASK-LLNL/BOOM} \\
\noalign{\vskip 2pt}
chemixhub-\allowbreak{}conductivity & CheMixHub & \cite{S:chemixhub2025} & not stated (code MIT; data licence not stated)$^{\dagger}$ & \url{https://github.com/chemcognition-lab/chemixhub} \\
\noalign{\vskip 1pt}
 & Polymer-electrolyte conductivity compilation (Bradford et al. 2023) & \cite{S:bradford2023} & MIT (data and code) & \url{https://github.com/learningmatter-mit/Chem-prop-pred} \\
\noalign{\vskip 1pt}
 & NIST ILThermo (SRD 147) ionic-liquid conductivity & \cite{S:ilthermo2007} & NIST Standard Reference Data (SRD 147); (c) U.S. Secretary of Commerce on behalf of the United States of America, all rights reserved & \url{https://ilthermo.boulder.nist.gov/} \\
\noalign{\vskip 2pt}
drugcomb-\allowbreak{}multiomics-\allowbreak{}synergy & DrugComb v1.4 & \cite{S:drugcomb2021} & CC BY 4.0 & \url{https://drugcomb.org/}; \url{https://zenodo.org/records/11102665} \\
\noalign{\vskip 1pt}
 & NCI-ALMANAC combination screen & \cite{S:almanac2017} & not stated (US federal NCI/DTP data; no explicit licence)$^{\dagger}$ & \url{https://wiki.nci.nih.gov/display/NCIDTPdata/NCI-ALMANAC} \\
\noalign{\vskip 1pt}
 & CellMiner / NCI-60 molecular profiles & \cite{S:cellminer2012} & not stated$^{\dagger}$ & \url{https://discover.nci.nih.gov/cellminer/} \\
\end{longtable}}

\clearpage
{\scriptsize
\setlength{\tabcolsep}{2pt}
\begin{longtable}{@{}>{\RaggedRight\arraybackslash}p{0.17\linewidth}
>{\centering\arraybackslash}p{0.065\linewidth}
>{\RaggedRight\arraybackslash}p{0.425\linewidth}
>{\RaggedRight\arraybackslash}p{0.23\linewidth}
>{\centering\arraybackslash}p{0.07\linewidth}@{}}
\caption{\textbf{Per-task input sources and modality families of \pro{}.}}
\label{tab:sources_modalities} \\
\multicolumn{5}{@{}p{0.97\linewidth}@{}}{``Sources'' counts the distinct input sources an agent receives for a task, under the definition used in the \base{} paper (each table column group, file or asset that carries independent information counts once; identifiers and the target do not). ``Modality families'' lists the families those sources belong to (molecule, sequence, structure, image, table, gene expression, protein abundance, other omics, spatial coordinates, text, time series). ``Multi-modal'' is the judgement under the same rubric: yes, at least two families carrying independent information; borderline, two families of which one is weakly informative; no, one family. Annotated from the public task descriptions and file headers on 2026-10-22 with the \base{} judgements as calibration; the per-task rationale is in \texttt{source\_modality.csv} in the code repository.} \\[2pt]
\toprule Task & Sources & Input sources & Modality families & Multi-modal \\ \midrule
\endfirsthead
\caption[]{\textbf{Per-task input sources and modality families (continued).}} \\
\toprule Task & Sources & Input sources & Modality families & Multi-modal \\ \midrule
\endhead
\midrule \multicolumn{5}{r}{\textit{continued on next page}} \\
\endfoot
\bottomrule
\endlastfoot
\multicolumn{5}{@{}l}{\textbf{Chemical biology (8 tasks)}} \\
boom-csd & 2 & SMILES molecular structure ; queried property (density or heat of formation) & 1: molecule & no \\
chemixhub-conductivity & 6 & component SMILES list ; component mole fractions ; temperature and pressure conditions ; component molecular weights ; conditions-only reference prediction ; electrolyte class (ionic liquid vs polymer electrolyte) & 2: molecule; table & yes \\
drugcomb-multiomics-synergy & 11 & drug A SMILES ; drug B SMILES ; drug protein-target annotations ; drug clinical phase ; cell-line and tissue metadata ; replicate count ; additive reference prediction (reference\_zip) ; single-agent monotherapy response table ; cell-line transcript expression ; cell-line SWATH proteomics ; cell-line miRNA profile & 5: molecule; gene expression; protein abundance; other omics: miRNA; table & yes \\
kano-molecular-property-prediction & 2 & SMILES molecular structure ; element-oriented knowledge graph and pre-computed element/functional-group embeddings & 1: molecule & no \\
massspecgym-msms-retrieval & 4 & MS/MS fragment spectrum (m/z and intensities) ; precursor m/z ; acquisition metadata (adduct, instrument type, collision energy) ; candidate SMILES & 3: mass spectra; molecule; table & yes \\
solubility & 1 & SMILES molecular structure & 1: molecule & no \\
tdc-dti-binding-affinity & 3 & drug SMILES ; target protein amino-acid sequence ; affinity-assay source dataset (kiba/davis/bindingdb) & 2: molecule; protein/peptide sequence & yes \\
\end{longtable}}

\bigskip
\begin{table}[!ht]
\caption{\textbf{The 22 agent configurations: score, failures, agent time and tokens.} Mean, mean normalized score over the 64 tasks ($\times 100$), failures scored 0; CI, 95\% percentile bootstrap over tasks (5,000 resamples). Time, agent wall-clock time summed over the 64 tasks; In and Out, input and output tokens in millions (input includes cached prompt tokens as reported by each API). \textdagger{}, two task scores taken from MLEvolve with Opus 5 because the API served empty completions (Methods). In Claude Code with Opus 5.5, two tasks were completed by Opus 5 through the CLI's refusal fallback (Methods). Repeat runs: the 64-task means of the three runs are 55.1, 54.1 and 54.4 for Codex CLI with GPT-6 Astra and 53.2, 53.6 and 53.8 for Claude Code with Opus 5.5; the table lists the first run. Three further leaderboard configurations (Codex CLI with GPT-6.1 Sol, 34.8; the minimal scaffold with DeepSeek V4 Pro, 33.7; and the minimal scaffold with Gemini 3.1 Pro, 33.5) fit neither design and are not analysed in the paper.}
\label{tab:configs}
{\footnotesize
\setlength{\tabcolsep}{2.5pt}
\begin{tabular*}{\linewidth}{@{\extracolsep{\fill}}llrrrrrr@{}}
\toprule
Scaffold & Backbone & Mean & 95\% CI & OK / fail & Time (h) & In (M) & Out (M) \\
\midrule
Codex CLI & GPT-6 Astra & 55.1 & 50.0--59.8 & 64 / 0 & 31.9 & 262.8 & 2.26 \\
Claude Code & Claude Opus 5.5 & 53.2 & 48.3--58.0 & 64 / 0 & 30.5 & 45.8 & 1.20 \\
OpenCode & GPT-6 Astra & 53.2 & 48.2--58.0 & 64 / 0 & 27.2 & 224.7 & 1.57 \\
OpenCode & Claude Opus 5.5 & 53.2 & 48.1--58.0 & 64 / 0 & 33.3 & 60.7 & 1.31 \\
Claude Code & Claude Opus 5 & 52.9 & 48.0--57.8 & 64 / 0 & 51.0 & 116.5 & 3.01 \\
MLEvolve & Claude Opus 5.5\textdagger{} & 52.7 & 47.7--57.6 & 64 / 0 & 127.0 & 272.4 & 30.41 \\
Biomni & Claude Opus 5.5 & 49.6 & 44.5--54.5 & 64 / 0 & 13.3 & 32.8 & 0.68 \\
Biomni & GPT-6 Astra & 48.4 & 43.5--53.2 & 64 / 0 & 12.6 & 8.0 & 0.82 \\
OpenCode & DeepSeek V4.1 Flash (1017) & 47.8 & 42.1--53.2 & 61 / 3 & 69.2 & 171.4 & 3.07 \\
OpenCode & Gemini 3.8 Flash & 47.8 & 42.6--52.6 & 63 / 1 & 33.8 & 197.1 & 3.41 \\
Biomni & Gemini 3.8 Flash & 47.2 & 42.3--52.2 & 63 / 1 & 29.8 & 108.3 & 2.72 \\
MLEvolve & GPT-6 Astra & 46.7 & 41.8--51.8 & 63 / 1 & 127.0 & 155.2 & 11.12 \\
Gemini CLI & Gemini 3.8 Flash & 46.7 & 41.3--52.0 & 62 / 2 & 44.7 & 279.2 & 4.64 \\
Minimal & Claude Opus 5.5 & 46.5 & 41.0--52.0 & 61 / 3 & 16.4 & 0.6 & 0.74 \\
Minimal & GPT-6 Astra & 46.5 & 41.6--51.4 & 63 / 1 & 11.2 & 0.3 & 0.27 \\
Codex CLI & GPT-5.6 Sol & 45.2 & 40.4--50.0 & 64 / 0 & 10.9 & 63.6 & 0.61 \\
MLEvolve & Gemini 3.8 Flash & 44.4 & 39.1--49.7 & 62 / 2 & 125.3 & 173.2 & 108.47 \\
Gemini CLI & Gemini 3.1 Pro (preview) & 41.4 & 36.1--46.7 & 61 / 3 & 28.7 & 104.6 & 1.38 \\
Biomni & DeepSeek V4.1 Flash (1017) & 40.6 & 35.0--46.4 & 58 / 6 & 35.3 & 115.8 & 2.75 \\
MLEvolve & DeepSeek V4.1 Flash (1017) & 38.9 & 33.1--44.9 & 55 / 9 & 126.3 & 141.9 & 72.19 \\
Minimal & Gemini 3.8 Flash & 37.4 & 32.1--42.9 & 59 / 5 & 19.9 & 0.5 & 1.49 \\
Minimal & DeepSeek V4.1 Flash (1017) & 33.6 & 27.9--39.4 & 59 / 5 & 17.3 & 0.5 & 2.06 \\
\bottomrule
\end{tabular*}}

\end{table}

\clearpage
\begin{table}[!htbp]
\caption{\textbf{Paired comparisons between configurations.} Mean difference in normalized score ($\times 100$) over the 64 tasks, number of tasks on which the first configuration is higher, tied or lower, and the two-sided Wilcoxon signed-rank $P$ value, raw and Holm-corrected over the 15 comparisons. A configuration can be higher on most tasks and still have a lower mean, because failed runs score 0.}\label{tab:paired}
\small\begin{tabular}{lrrrrrr}\toprule
Comparison & Mean diff. & Higher & Tied & Lower & $P$ & $P_{\mathrm{Holm}}$ \\\midrule
Codex CLI vs OpenCode, GPT-6 Astra & +1.88 & 51 & 0 & 13 & $<$0.0001 & $<$0.0001 \\
Claude Code vs OpenCode, Opus 5.5 & +0.04 & 34 & 0 & 30 & 0.2326 & 0.6978 \\
Gemini CLI vs OpenCode, Gemini 3.8 Flash & -1.06 & 37 & 1 & 26 & 0.4017 & 0.8033 \\
OpenCode vs Minimal, GPT-6 Astra & +6.70 & 52 & 0 & 12 & $<$0.0001 & $<$0.0001 \\
OpenCode vs Minimal, Opus 5.5 & +6.68 & 51 & 0 & 13 & $<$0.0001 & $<$0.0001 \\
OpenCode vs Minimal, Gemini 3.8 Flash & +10.32 & 53 & 1 & 10 & $<$0.0001 & $<$0.0001 \\
OpenCode vs Minimal, DeepSeek V4.1 Flash & +14.16 & 54 & 1 & 9 & $<$0.0001 & $<$0.0001 \\
GPT-6 Astra vs GPT-5.6 Sol, Codex CLI & +9.87 & 59 & 0 & 5 & $<$0.0001 & $<$0.0001 \\
Opus 5.5 vs Opus 5, Claude Code & +0.27 & 39 & 0 & 25 & 0.1107 & 0.4429 \\
Gemini 3.8 Flash vs Gemini 3.1 Pro, Gemini CLI & +5.28 & 49 & 0 & 15 & $<$0.0001 & 0.0006 \\
MLEvolve vs Codex CLI, GPT-6 Astra & -8.36 & 7 & 0 & 57 & $<$0.0001 & $<$0.0001 \\
MLEvolve vs Claude Code, Opus 5.5 & -0.49 & 30 & 0 & 34 & 0.4867 & 0.8033 \\
MLEvolve vs OpenCode, Gemini 3.8 Flash & -3.40 & 26 & 0 & 38 & 0.0375 & 0.1877 \\
MLEvolve vs OpenCode, DeepSeek V4.1 Flash & -8.86 & 19 & 0 & 45 & 0.0020 & 0.0143 \\
Codex CLI + GPT-6 Astra vs Claude Code + Opus 5.5 & +1.85 & 44 & 0 & 20 & 0.0035 & 0.0208 \\
\bottomrule\end{tabular}
\end{table}

\clearpage
{\scriptsize
\setlength{\tabcolsep}{2pt}
\setlength{\emergencystretch}{1em}
\begin{longtable}{@{}>{\RaggedRight\arraybackslash}p{0.15\linewidth}
>{\RaggedRight\arraybackslash}p{0.49\linewidth}
>{\RaggedRight\arraybackslash}p{0.24\linewidth}
>{\RaggedLeft\arraybackslash}p{0.05\linewidth}@{}}
\caption{\textbf{Primary metric, normalization and chance level of each task.}}
\label{tab:metrics} \\
\multicolumn{4}{@{}p{0.955\linewidth}@{}}{Definitions follow the grading code of the release. Every task score is finally clipped to $[0,1]$ and reported $\times 100$; an invalid submission (schema, identifier or value errors) scores 0 for the task, and 100 is the formal maximum of every metric (no noise ceiling was measured, so it need not be reachable in practice). ``Chance'' is the score of an uninformed submission on the 0--100 scale, measured with the released graders (constant or reference prediction for regression and ranking scores; uniformly random labels, random rankings or random scores otherwise; mean of 3--5 seeds). The tasks fall into two families. (i) Chance-zero metrics (36 tasks): correlations reported as $\mathrm{clip}(r,0,1)$, calibrated regression scores, $R^{2}$, average precision or F1 expressed as a gain over label prevalence, skill scores and scores offset against a public reference prediction, together with exact-match and segmentation scores that have no chance agreement. (ii) Metrics kept on their native scale (28 tasks): ROC-AUC and the concordance index (chance 50), and macro-F1, accuracy, average precision, reciprocal rank, Fmax, sequence recovery and sensitivity at fixed specificity, whose chance level depends on the number of classes or candidates and is listed per task. $r$, Pearson correlation; $\rho$, Spearman correlation; AP, average precision; $p$, positive prevalence in the test set; SD, standard deviation; n.d., not defined. \textsuperscript{a}An input-independent training term-frequency prior scores 39.6 on this task. \textsuperscript{b}Exact counts have no natural chance level; the all-zero template scores 0. \textsuperscript{c}The training climatology scores 0; a uniform forecast scores 14.8.} \\[2pt]
\toprule Task & Primary metric and aggregation & Normalization to 0--100 & Chance \\ \midrule
\endfirsthead
\caption[]{\textbf{Primary metric, normalization and chance level of each task (continued).}} \\
\toprule Task & Primary metric and aggregation & Normalization to 0--100 & Chance \\ \midrule
\endhead
\midrule \multicolumn{4}{r}{\textit{continued on next page}} \\
\endfoot
\bottomrule
\endlastfoot
\multicolumn{4}{@{}l}{\textbf{Chemical biology (8 tasks)}} \\
boom-\allowbreak{}csd & Calibrated regression score: 0.5\,$\times$\,clip(Pearson r,0,1) + 0.5\,$\times$\,clip(1 - RMSE/\allowbreak{}SD(y\_\allowbreak{}true),0,1). \textit{Aggregation:} computed in each of 4 cells (source density/\allowbreak{}hof $\times$ low/\allowbreak{}high extrapolation tail); equal-weight mean of the 4 cells. & Components clipped inside the grader & 0 \\
chemixhub-\allowbreak{}conductivity & Calibrated regression score on residuals after subtracting the public conditions-only reference (test.csv reference\_\allowbreak{}ln\_\allowbreak{}conductivity) from truth and prediction: 0.5\,$\times$\,clip(Pearson r,0,1) + 0.5\,$\times$\,clip(1 - RMSE/\allowbreak{}SD(y\_\allowbreak{}true - ref),0,1). \textit{Aggregation:} computed within each of the 2 sources; equal-weight mean. & Offset against reference prediction, components clipped inside the grader & 0 \\
drugcomb-\allowbreak{}multiomics-\allowbreak{}synergy & Partial Pearson correlation: truth and prediction projected onto the orthogonal complement of span\{cell-line indicators, public additive reference\_\allowbreak{}zip\}, then Pearson r of the projections. \textit{Aggregation:} computed within each of 9 tissues ($\geq$ 20 rows); mean over tissues. & Clip(mean r, 0, 1) in the grader (and again by the evaluator) & 0 \\
kano-\allowbreak{}molecular-\allowbreak{}property-\allowbreak{}prediction & Classification instances: average precision per scored target minus prevalence, divided by (1 - prevalence); regression instances: R$^{2}$ = 1 - MSE/\allowbreak{}Var(y) per target. \textit{Aggregation:} per-target values clipped to [0,1], mean over scored targets of an instance, then equal-weight mean over 12 instances. & Gain over prevalence (classification) /\allowbreak{} R$^{2}$ (regression), each clipped to [0,1] per target & 0 \\
massspecgym-\allowbreak{}msms-\allowbreak{}retrieval & Mean reciprocal rank of the single correct candidate structure; ties broken uniformly at random (expected reciprocal rank). \textit{Aggregation:} per query (2,996 queries, 13-256 candidates each); mean over queries. & None (native scale) & 3.1 \\
solubility & Calibrated regression score: 0.5\,$\times$\,clip(Pearson r,0,1) + 0.5\,$\times$\,clip(1 - RMSE/\allowbreak{}SD(y\_\allowbreak{}true),0,1). \textit{Aggregation:} pooled over all test molecules (no grouping). & Components clipped inside the grader & 0 \\
tdc-\allowbreak{}dti-\allowbreak{}binding-\allowbreak{}affinity & Calibrated regression score: 0.5\,$\times$\,clip(Pearson r,0,1) + 0.5\,$\times$\,clip(1 - RMSE/\allowbreak{}SD(y\_\allowbreak{}true),0,1). \textit{Aggregation:} computed within each of 3 sources; fixed-weight mean 0.4 KIBA + 0.4 BindingDB + 0.2 DAVIS. & Components clipped inside the grader & 0 \\
uspto-\allowbreak{}reaction-\allowbreak{}product & Top-1 exact-match accuracy: first candidate's cleaned product fragment set (RDKit canonical, atom maps removed, single-heavy-atom fragments dropped) equals the reference set. \textit{Aggregation:} fraction over all test reactions (1 instance). & None (native scale) & 0 \\
\midrule
\multicolumn{4}{@{}l}{\textbf{Imaging (8 tasks)}} \\
calcium-\allowbreak{}neuron-\allowbreak{}segmentation & real: neuron mask F1 = 2\,$\times$\,matched/\allowbreak{}(n\_\allowbreak{}ref + n\_\allowbreak{}pred), one-to-one greedy matching at IoU $>$ 0.5; simulated: trace-weighted F1 = 2\,$\times$\,sum over matched pairs of max(0, Pearson(pred trace, ref trace))/\allowbreak{}(n\_\allowbreak{}ref + n\_\allowbreak{}pred). \textit{Aggregation:} each pooled (micro) over all crops/\allowbreak{}substacks of the instance; equal-weight mean of the 2 instances. & Per-instance clip to [0,1] & 0 \\
cryocrab-\allowbreak{}quality & Pearson correlation between predicted and measured CTF-fit resolution. \textit{Aggregation:} within each test session (EMPIAR entry) with $\geq$ 6 micrographs (57 of 58); mean over sessions. & Clip(mean r, 0, 1) by the evaluator & 0 \\
drugclass-\allowbreak{}if & Macro-F1 over the 3 fixed treatment classes. \textit{Aggregation:} pooled over all test fields of view. & None (native scale) & 33.4 \\
mammovqa-\allowbreak{}external & Macro-F1 over the union of true and predicted answer classes. \textit{Aggregation:} within each of 4 question topics; equal-weight mean over topics (12 mislabelled rows required but not scored). & None (native scale) & 21.5 \\
medmnist3d-\allowbreak{}decathlon & Macro-F1 over the union of true and predicted classes. \textit{Aggregation:} within each of 5 sources (2-11 classes each); equal-weight mean over sources. & None (native scale) & 35.1 \\
microscopy-\allowbreak{}instance-\allowbreak{}segmentation & Mean segmentation accuracy (mSA): mean over IoU thresholds 0.50..0.95 (step 0.05) of TP/\allowbreak{}(TP+FP+FN), TP = maximum one-to-one matching at IoU $\geq$ t. \textit{Aggregation:} per image; mean over images within an instance; equal-weight mean over 2 instances (covid\_\allowbreak{}if, deepbacs). & None (native scale) & 0 \\
rocov2-\allowbreak{}retrieval & Average precision of the candidate scores (sklearn average\_\allowbreak{}precision\_\allowbreak{}score); one correct caption among 256 per query, so it equals the reciprocal rank when there are no ties. \textit{Aggregation:} per query (1,588 queries); mean over queries. & None (native scale) & 2.4 \\
subcell-\allowbreak{}localization & 0.5 \,$\times$\, macro-F1 at threshold 0.5 + 0.5 \,$\times$\, macro average precision. \textit{Aggregation:} macro over the 14 scored labels (3 of 17 labels with $<$ 5 positive test proteins excluded), computed on crops pooled. & None (native scale) & 11.6 \\
\midrule
\multicolumn{4}{@{}l}{\textbf{Network biology (8 tasks)}} \\
causalbench-\allowbreak{}grn & ROC-AUC over the strength-matched scored rows (private mask). \textit{Aggregation:} within each (cell line, downstream gene) group (2,252 scored groups); mean over groups. & None (native scale) & 50 \\
lnctard-\allowbreak{}relation & Average precision over a query's candidates (sklearn average\_\allowbreak{}precision\_\allowbreak{}score). \textit{Aggregation:} per scored query (547; cold-start queries excluded by the private scored flag); mean over queries. & None (native scale) & 19.2 \\
music-\allowbreak{}seccomplex & Prevalence-normalized average precision: (AP - p)/\allowbreak{}(1 - p), p = fraction of positive pairs (0.0041). \textit{Aggregation:} pooled over all 147,153 test pairs. & Gain over prevalence, floored at 0 in the grader & 0 \\
ppi-\allowbreak{}cross-\allowbreak{}species & Average precision (AUPR) of the interaction scores; for the 6 pair instances the logit scores are first two-way centred per protein on the test pair graph. \textit{Aggregation:} per instance; equal-weight mean over 7 instances (6 pair instances, mutation\_\allowbreak{}effect). & Per-instance clip to [0,1]; no prevalence correction & 24.2 \\
pring-\allowbreak{}cross-\allowbreak{}species-\allowbreak{}ppi & ROC-AUC after rank-to-logit transform and two-way per-protein centring of the scores on the test pair graph. \textit{Aggregation:} within each of 4 species; equal-weight mean. & None (native scale) & 50 \\
scigym-\allowbreak{}sbml & Exact integer accuracy: fraction of rows whose predicted count equals the true count. \textit{Aggregation:} per target column; mean over the 3 targets (deleted species, transitions, species glyphs). & None (native scale) & n.d.\textsuperscript{b} \\
swissprot-\allowbreak{}protein-\allowbreak{}function & Protein-centric Fmax: maximum over thresholds 0.00..1.00 of the harmonic mean of mean precision and mean recall over proteins, on GO terms of the instance vocabulary. \textit{Aggregation:} per instance; weighted mean, 0.8/\allowbreak{}3 for each Swiss-Prot instance (MF, BP, CC) and 0.1 for each neXtProt instance (MF, BP). & Per-instance clip to [0,1]; native Fmax scale & 0.5\textsuperscript{a} \\
synlethdb-\allowbreak{}sl & Average precision over a query gene's candidate partners (sklearn average\_\allowbreak{}precision\_\allowbreak{}score). \textit{Aggregation:} per query (381 queries, about 138 candidates and 4.3 positives each); mean over queries. & None (native scale) & 8.2 \\
\midrule
\multicolumn{4}{@{}l}{\textbf{Perturbation (8 tasks)}} \\
cellverse-\allowbreak{}cta & Macro-F1 over the union of true and predicted labels (21 allowed labels incl. Novel\_\allowbreak{}unassigned; all 21 occur in the answers). \textit{Aggregation:} pooled over all test nuclei. & None (native scale) & 4.2 \\
cpjump-\allowbreak{}genetic-\allowbreak{}matching & ROC-AUC of the annotated target gene among 65 candidate genes. \textit{Aggregation:} per test compound (130); mean over compounds. & None (native scale) & 50 \\
gdsc-\allowbreak{}lineage-\allowbreak{}holdout-\allowbreak{}response & Pearson correlation between predicted and measured selective response across cell lines. \textit{Aggregation:} within each (held-out lineage, compound) group with $\geq$ 10 lines; mean over compounds within a lineage, then over 5 lineages. & Clip(mean r, 0, 1) by the evaluator & 0 \\
perturbench-\allowbreak{}response & Pearson correlation of the context $\times$ treatment interaction: (y - context baseline - treatment reference) projected onto the orthogonal complement of span\{1, baseline, reference\} on the source's scoring genes. \textit{Aggregation:} per test row; mean over rows within each of 4 sources; equal-weight mean over sources. & Offset against public references + projection; clip(mean r, 0, 1) by the evaluator & 0 \\
scgenescope-\allowbreak{}treatment & Macro-F1 over the 29 treatment classes (union of true and predicted labels). \textit{Aggregation:} pooled over all test cells. & None (native scale) & 3.3 \\
sciplex-\allowbreak{}distribution & Per condition, after subtracting the training-only context baseline: 0.5\,$\times$\,Pearson(mean shift, top-50 DEGs) + 0.25\,$\times$\,Pearson(log-SD ratio, 1,000 genes) + 0.25\,$\times$\,Pearson(median shift, 1,000 genes). \textit{Aggregation:} mean s over 278 responsive test conditions. & Offset against the all-zero prediction: (s - $|$s0$|$)/\allowbreak{}(1 - $|$s0$|$), s0 = score of an all-zero prediction on the same answers; then clip to [0,1] by the evaluator & 0 \\
tahoe-\allowbreak{}single-\allowbreak{}cell & Pearson correlation between flattened truth and prediction residual matrices (held-out lines $\times$ 2,000 genes), residual = response - public treatment-by-plate shift, centred within each plate $\times$ cell-line context. \textit{Aggregation:} per test treatment (631); mean s over treatments. & Offset against the all-zero prediction: (s - $|$s0$|$)/\allowbreak{}(1 - $|$s0$|$); then clip to [0,1] by the evaluator & 0 \\
xatlas-\allowbreak{}orion-\allowbreak{}perturb & Pearson correlation of (y - line baseline) after projecting out span\{1, baseline, line residual axis\}, on the line's scoring genes minus private cis/\allowbreak{}target exclusions. \textit{Aggregation:} per test row; mean over rows within each of 2 cell lines; equal-weight mean over lines. & Offset against public baseline + projection; clip(mean r, 0, 1) by the evaluator & 0 \\
\midrule
\multicolumn{4}{@{}l}{\textbf{Phenotype--disease (8 tasks)}} \\
clinvar-\allowbreak{}disease-\allowbreak{}variant-\allowbreak{}effect & Average precision minus positive prevalence, divided by (1 - prevalence). \textit{Aggregation:} per instance; per-instance clip to [0,1]; equal-weight mean over 6 instances. & Gain over prevalence, clipped to [0,1] per instance & 0 \\
covid-\allowbreak{}hospitalization-\allowbreak{}forecast & Ranked probability skill score: 1 - (RPS\_\allowbreak{}t1 + RPS\_\allowbreak{}t3)/\allowbreak{}(RPS\_\allowbreak{}clim\_\allowbreak{}t1 + RPS\_\allowbreak{}clim\_\allowbreak{}t3), RPS = sum over 4 cumulative thresholds of squared error of the 5-class ordinal forecast, climatology = fixed training class frequencies. \textit{Aggregation:} RPS averaged over 50 states within each of 2 forecast origins, then over origins, per horizon (1 and 3 weeks). & Skill score against training climatology, clipped to [0,1] & 0\textsuperscript{c} \\
cptac-\allowbreak{}pan-\allowbreak{}protein-\allowbreak{}imputation & 0.5 \,$\times$\, mean over patients of Pearson r across observed proteins + 0.5 \,$\times$\, mean over proteins of Pearson r across patients (observed entries only; private mask). \textit{Aggregation:} per test cohort (8); equal-weight mean over cohorts. & Clip(mean r, 0, 1) by the evaluator & 0 \\
gutcrc-\allowbreak{}cross-\allowbreak{}cohort-\allowbreak{}detection & Sensitivity for (mainly early-stage) cancer at 90\% specificity: ROC curve interpolated at FPR = 0.10; unscored cancers flagged -1 in the answers. \textit{Aggregation:} within each of 6 test cohorts ($\geq$ 20 rows, $\geq$ 5 scored positives and negatives); equal-weight mean. & None (native scale); not rescaled to chance & 10 \\
msk-\allowbreak{}met-\allowbreak{}tumor-\allowbreak{}origin & Macro-F1 over the union of true and predicted labels (22 cancer types). \textit{Aggregation:} pooled over all test patients. & None (native scale) & 3.4 \\
survpath-\allowbreak{}multimodal & Harrell concordance index of the risk score (ties in risk count 0.5; tied times not comparable). \textit{Aggregation:} within each of 5 cancer cohorts; mean weighted by the number of observed deaths. & None (native scale) & 50 \\
target-\allowbreak{}aml-\allowbreak{}cytogenetics & Macro-F1 over 7 fixed cytogenetic classes. \textit{Aggregation:} pooled over all test patients. & None (native scale) & 13 \\
tcga-\allowbreak{}pancan-\allowbreak{}multiomics-\allowbreak{}survival & Harrell concordance index of the risk score (ties in risk count 0.5; tied times not comparable). \textit{Aggregation:} within each of 16 cohorts; mean weighted by the number of observed deaths. & None (native scale) & 50 \\
\midrule
\multicolumn{4}{@{}l}{\textbf{Sequence and genomics (8 tasks)}} \\
aso-\allowbreak{}knockdown-\allowbreak{}efficacy & Pearson correlation between predicted and measured inhibition after centring both within each (source, chemistry pattern) group. \textit{Aggregation:} pooled within each of 2 sources; equal-weight mean over sources. & Clip(mean r, 0, 1) by the evaluator & 0 \\
cgbench-\allowbreak{}variant & Macro-F1 over the 5 assertion classes (union of true and predicted labels). \textit{Aggregation:} pooled over all test variants. & None (native scale) & 19.1 \\
deepstarr-\allowbreak{}genomic-\allowbreak{}sequence & Matthews correlation coefficient (13 classification instances; multiclass MCC for two 3-class instances) or Pearson r of developmental enhancer activity (deepstarr instance). \textit{Aggregation:} per instance; per-instance clip to [0,1]; equal-weight mean over 14 instances. & Clip(x, 0, 1) per instance & 0 \\
de-\allowbreak{}novo-\allowbreak{}peptide-\allowbreak{}sequencing-\allowbreak{}from-\allowbreak{}tandem-\allowbreak{}mass-\allowbreak{}spectrometry & 0.5 \,$\times$\, peptide accuracy (all residues mass-matched, same length) + 0.5 \,$\times$\, amino-acid F1 (Casanovo prefix/\allowbreak{}suffix mass matching, 0.5 Da cumulative /\allowbreak{} 0.1 Da residue tolerance). \textit{Aggregation:} counts pooled over the spectra of an instance; equal-weight mean over 5 instances. & None (native scale) & 0 \\
escape-\allowbreak{}amp-\allowbreak{}multilabel & 0.5 \,$\times$\, mean AP gain + 0.5 \,$\times$\, mean F1 gain; AP gain = (AP - p)/\allowbreak{}(1 - p); F1 gain = (F1 at 0.5 - f0)/\allowbreak{}(1 - f0) with f0 = 2p/\allowbreak{}(1+p) (F1 of predicting all positive), p = label prevalence. \textit{Aggregation:} per label (5 labels with $\geq$ 30 positives and negatives); each gain clipped to [0,1]; mean over labels. & Gain over prevalence (AP) and over the all-positive F1 floor, clipped per label & 0 \\
proteingym-\allowbreak{}fitness & Spearman correlation between predicted and measured fitness. \textit{Aggregation:} within each of 206 DMS assays; equal-weight mean over assays. & Clip(mean rho, 0, 1) by the evaluator & 0 \\
sirna-\allowbreak{}knockdown-\allowbreak{}efficacy & Pearson correlation between predicted and measured inhibition after centring both within each source (assay). \textit{Aggregation:} pooled over all 1,531 test duplexes of the 4 sources (single evaluation group). & Clip(r, 0, 1) by the evaluator & 0 \\
yeast-\allowbreak{}promoter-\allowbreak{}expression & Weighted mean of max(r, 0)$^{2}$ over 9 test subsets (DREAM 2022 weights, total 4.65); r = Pearson on expression for 6 single-sequence subsets and on expression differences for 3 paired subsets (SNV, motif perturbation, motif tiling). \textit{Aggregation:} per subset; weighted mean (1 instance); former public-leaderboard sequences dropped. & None (native scale) & 0 \\
\midrule
\multicolumn{4}{@{}l}{\textbf{Single-cell and spatial (8 tasks)}} \\
dlpfc-\allowbreak{}spatial-\allowbreak{}domain & Macro-F1 over the cortical layers present in the section's reference annotation. \textit{Aggregation:} within each of 8 test sections; equal-weight mean over sections. & None (native scale) & 14 \\
gastrulation-\allowbreak{}kinetic-\allowbreak{}state & 8/\allowbreak{}42 \,$\times$\, stage accuracy (per embryo: modal predicted stage of its cells equals the true stage) + 34/\allowbreak{}42 \,$\times$\, macro-F1 of cell fate over the 34 fates present in the answers. \textit{Aggregation:} stage: fraction of 13 test embryos; fate: per cell, pooled. & None (native scale) & 3.8 \\
hest-\allowbreak{}coad-\allowbreak{}spatial-\allowbreak{}expression & 0.5 \,$\times$\, mean over spots of Pearson r across 50 genes after centring each gene by its section mean + 0.5 \,$\times$\, mean over genes of Pearson r across spots. \textit{Aggregation:} per test section; mean over sections. & Clip(mean r, 0, 1) by the evaluator & 0 \\
lung-\allowbreak{}cite-\allowbreak{}cell-\allowbreak{}matching & Top-1 matching accuracy: submitted ADT cell id equals the paired cell. \textit{Aggregation:} fraction over all test RNA cells (128 candidates each). & None (native scale) & 0.8 \\
openproblems-\allowbreak{}rna-\allowbreak{}atac & 0.5 \,$\times$\, mean over metacells of Pearson r across peaks + 0.5 \,$\times$\, mean over peaks of Pearson r across metacells (targets are centred on the training mean). \textit{Aggregation:} per test donor (2); equal-weight mean. & Clip(mean r, 0, 1) by the evaluator & 0 \\
openproblems-\allowbreak{}rna-\allowbreak{}protein & 0.5 \,$\times$\, mean over cells of Pearson r across 140 proteins + 0.5 \,$\times$\, mean over proteins of Pearson r across cells (targets are centred on the training mean). \textit{Aggregation:} per test group (6 donor $\times$ day groups); equal-weight mean. & Clip(mean r, 0, 1) by the evaluator & 0 \\
seqfish-\allowbreak{}deconvolution & Coefficient of determination R$^{2}$ = 1 - MSE/\allowbreak{}Var(y) of the predicted cell-type proportion. \textit{Aggregation:} per cell type across test spots, clipped to [0,1]; mean over 16 cell types. & Clip to [0,1] per cell type & 0 \\
tabula-\allowbreak{}sapiens-\allowbreak{}ood-\allowbreak{}annotation & Macro-F1 over the labels present in the block's truth or predictions (19-class vocabulary). \textit{Aggregation:} within each donor $\times$ organ block; mean over blocks within a donor; equal-weight mean over 20 donors. & None (native scale) & 2.3 \\
\midrule
\multicolumn{4}{@{}l}{\textbf{Structure (8 tasks)}} \\
asep-\allowbreak{}epitope & Prevalence-normalized average precision: (AP\_\allowbreak{}c - p\_\allowbreak{}c)/\allowbreak{}(1 - p\_\allowbreak{}c), p\_\allowbreak{}c = epitope fraction of the complex's residues. \textit{Aggregation:} per complex, floored at 0; equal-weight mean over 73 complexes. & Gain over prevalence, floored at 0 per complex & 0 \\
cath-\allowbreak{}inverse-\allowbreak{}folding & Sequence recovery: fraction of positions where the predicted residue equals the native one. \textit{Aggregation:} per chain; median over all chains of an instance; equal-weight mean over 2 instances (cath, pdb2022). & Percent divided by 100, clipped to [0,1] & 5.1 \\
cpsea-\allowbreak{}cyclic-\allowbreak{}peptide-\allowbreak{}affinity & Calibrated regression score: 0.5\,$\times$\,clip(Pearson r,0,1) + 0.5\,$\times$\,clip(1 - RMSE/\allowbreak{}SD(y\_\allowbreak{}true),0,1). \textit{Aggregation:} per output (Rosetta dG, Vina score), pooled over test complexes; mean of the 2 outputs. & Components clipped inside the grader & 0 \\
decoydb-\allowbreak{}pose-\allowbreak{}classify & Spearman correlation between predicted and true ligand RMSD of docking poses. \textit{Aggregation:} within each of 196 test complexes (in-site poses only); mean over complexes. & Clip(mean rho, 0, 1) by the evaluator & 0 \\
denoise-\allowbreak{}cryo & 0.5 \,$\times$\, backbone F1 (mean over 5 atom types of one-to-one F1 with one-voxel tolerance) + 0.5 \,$\times$\, residue-identity F1 (C-alpha position within one voxel and correct amino acid). \textit{Aggregation:} per test volume; mean over volumes within a sequence cluster; mean over clusters. & None (native scale) & 0 \\
ppi-\allowbreak{}interface-\allowbreak{}residues & ROC-AUC of per-residue interface scores. \textit{Aggregation:} per chain (799 of 800 chains with $\geq$ 5 interface and $\geq$ 5 non-interface residues); unweighted mean over chains. & None (native scale) & 50 \\
psbench-\allowbreak{}multimer-\allowbreak{}mqa & Mean of two clipped components: A = mean over 193 pools of Pearson r(predicted, true TM-score) across the pool's 15 models; B = mean over pipeline groups (true TM span $\geq$ 0.02) of Pearson r across the pipeline's 5 models. \textit{Aggregation:} A per pool, B per (pool, pipeline) group; score = (clip(A,0,1) + clip(B,0,1))/\allowbreak{}2. & Each component clip(mean r, 0, 1) inside the grader & 0 \\
skempi-\allowbreak{}ppi-\allowbreak{}ddg & Spearman correlation between predicted and measured ddG. \textit{Aggregation:} within each test complex with $\geq$ 5 mutants and non-zero SD of ddG (40 of 74); unweighted mean over complexes. & Clip(mean rho, 0, 1) by the evaluator & 0 \\

\end{longtable}}

\clearpage
{\scriptsize
\setlength{\tabcolsep}{4pt}
\begin{longtable}{@{}>{\RaggedRight\arraybackslash}p{0.27\linewidth}cl rrrr >{\RaggedRight\arraybackslash}p{0.27\linewidth}@{}}
\caption{\textbf{Per-task scores of the reproduced published-method reference and of the agents.}}
\label{tab:pertask} \\
\multicolumn{8}{@{}p{0.98\linewidth}@{}}{All scores $\times 100$. Tier 1, method published for the task's dataset; tier 2, strongest published method of the same problem class. Group, comparability of our test set with the original study: close (same data and metric family), rebuilt (new split, removed inputs or new metric) or mixed (the grade differs between instances or applies to part of the task). The nine tier-1 tasks in the close group ($\ast$) form the published-method baseline. Published, the primary reproduced published method (named in Supplementary Table 1); Codex, Codex CLI with GPT-6 Astra; Claude, Claude Code with Opus 5.5. Best, the highest score among the 22 configurations and the configuration that reached it. The scores of all 22 configurations on every task are in the leaderboard files of the code repository.} \\
\toprule Task & Tier & Group & Published & Codex & Claude & Best & Best configuration \\ \midrule
\endfirsthead
\caption[]{\textbf{Per-task scores of the reproduced published-method reference and of the agents (continued).}} \\
\toprule Task & Tier & Group & Published & Codex & Claude & Best & Best configuration \\ \midrule
\endhead
\midrule \multicolumn{8}{r}{\textit{continued on next page}} \\
\endfoot
\bottomrule
\endlastfoot
\multicolumn{8}{@{}l}{\textbf{Chemical biology (8 tasks)}} \\
boom-\allowbreak{}csd & 1 & mixed & 77.3 & 86.8 & 83.8 & 88.0 & Claude Code + Opus 5 \\
chemixhub-\allowbreak{}conductivity & 1 & rebuilt & 54.8 & 56.7 & 56.8 & 57.8 & MLEvolve + Gemini 3.8 Flash \\
drugcomb-\allowbreak{}multiomics-\allowbreak{}synergy & 2 & rebuilt & 28.5 & 47.6 & 45.7 & 48.6 & MLEvolve + Opus 5.5 \\
kano-\allowbreak{}molecular-\allowbreak{}property-\allowbreak{}prediction & 1 & rebuilt & 68.6 & 69.2 & 69.7 & 69.7 & Claude Code + Opus 5.5 \\
massspecgym-\allowbreak{}msms-\allowbreak{}retrieval & 2 & rebuilt & 67.8 & 77.0 & 79.9 & 80.6 & MLEvolve + Opus 5.5 \\
solubility & 2 & mixed & 72.4 & 74.6 & 74.8 & 75.1 & OpenCode + DeepSeek V4.1 Flash \\
tdc-\allowbreak{}dti-\allowbreak{}binding-\allowbreak{}affinity & 2 & rebuilt & 20.4 & 29.6 & 33.0 & 33.0 & Claude Code + Opus 5.5 \\
uspto-\allowbreak{}reaction-\allowbreak{}product$^{\ast}$ & 1 & close & 84.2 & 84.1 & 81.2 & 84.1 & Codex CLI + GPT-6 Astra \\
\multicolumn{8}{@{}l}{\textbf{Imaging (8 tasks)}} \\
calcium-\allowbreak{}neuron-\allowbreak{}segmentation & 2 & rebuilt & 72.8 & 83.2 & 82.5 & 83.2 & Codex CLI + GPT-6 Astra \\
cryocrab-\allowbreak{}quality & 2 & rebuilt & 37.7 & 49.0 & 43.1 & 49.0 & Codex CLI + GPT-6 Astra \\
drugclass-\allowbreak{}if & 1 & rebuilt & 78.5 & 88.2 & 83.8 & 88.9 & MLEvolve + Opus 5.5 \\
mammovqa-\allowbreak{}external & 1 & rebuilt & 46.4 & 51.2 & 54.4 & 55.2 & Claude Code + Opus 5 \\
medmnist3d-\allowbreak{}decathlon & 1 & rebuilt & 57.5 & 76.6 & 73.0 & 76.9 & OpenCode + DeepSeek V4.1 Flash \\
microscopy-\allowbreak{}instance-\allowbreak{}segmentation & 2 & mixed & 56.9 & 55.7 & 59.1 & 59.1 & Claude Code + Opus 5.5 \\
rocov2-\allowbreak{}retrieval & 2 & mixed & 53.1 & 56.1 & 54.1 & 57.5 & MLEvolve + Opus 5.5 \\
subcell-\allowbreak{}localization & 2 & rebuilt & 76.7 & 82.8 & 87.1 & 87.5 & MLEvolve + Opus 5.5 \\
\multicolumn{8}{@{}l}{\textbf{Network biology (8 tasks)}} \\
causalbench-\allowbreak{}grn & 2 & rebuilt & 56.4 & 68.1 & 67.2 & 68.1 & Codex CLI + GPT-6 Astra \\
lnctard-\allowbreak{}relation & 2 & rebuilt & 24.9 & 38.0 & 36.5 & 38.0 & Codex CLI + GPT-6 Astra \\
music-\allowbreak{}seccomplex & 2 & rebuilt & 5.9 & 10.5 & 10.0 & 12.1 & MLEvolve + Opus 5.5 \\
ppi-\allowbreak{}cross-\allowbreak{}species & 1 & rebuilt & 37.0 & 36.8 & 41.6 & 41.6 & Claude Code + Opus 5.5 \\
pring-\allowbreak{}cross-\allowbreak{}species-\allowbreak{}ppi & 1 & rebuilt & 73.7 & 69.4 & 79.8 & 82.6 & MLEvolve + Opus 5.5 \\
scigym-\allowbreak{}sbml & 2 & rebuilt & 43.7 & 57.8 & 55.1 & 57.8 & Codex CLI + GPT-6 Astra \\
swissprot-\allowbreak{}protein-\allowbreak{}function & 2 & mixed & 53.0 & 53.3 & 51.6 & 53.5 & OpenCode + GPT-6 Astra \\
synlethdb-\allowbreak{}sl & 1 & rebuilt & 16.8 & 27.0 & 27.7 & 28.6 & MLEvolve + Opus 5.5 \\
\multicolumn{8}{@{}l}{\textbf{Perturbation (8 tasks)}} \\
scgenescope-\allowbreak{}treatment$^{\ast}$ & 1 & close & 26.2 & 31.4 & 31.1 & 33.2 & MLEvolve + Opus 5.5 \\
cpjump-\allowbreak{}genetic-\allowbreak{}matching & 1 & rebuilt & 68.3 & 69.3 & 72.3 & 72.3 & Claude Code + Opus 5.5 \\
gdsc-\allowbreak{}lineage-\allowbreak{}holdout-\allowbreak{}response & 2 & rebuilt & 16.0 & 17.3 & 16.9 & 18.1 & Biomni + Gemini 3.8 Flash \\
cellverse-\allowbreak{}cta & 2 & rebuilt & 47.4 & 66.3 & 66.0 & 66.5 & MLEvolve + Opus 5.5 \\
perturbench-\allowbreak{}response & 1 & rebuilt & 30.5 & 37.4 & 37.4 & 39.0 & OpenCode + Opus 5.5 \\
sciplex-\allowbreak{}distribution & 2 & rebuilt & 41.8 & 53.0 & 44.9 & 55.2 & MLEvolve + Opus 5.5 \\
tahoe-\allowbreak{}single-\allowbreak{}cell & 2 & rebuilt & 30.9 & 53.8 & 53.6 & 54.3 & Claude Code + Opus 5 \\
xatlas-\allowbreak{}orion-\allowbreak{}perturb & 2 & rebuilt & 7.0 & 7.5 & 7.8 & 8.2 & MLEvolve + Opus 5.5 \\
\multicolumn{8}{@{}l}{\textbf{Phenotype--disease (8 tasks)}} \\
clinvar-\allowbreak{}disease-\allowbreak{}variant-\allowbreak{}effect & 1 & mixed & 76.2 & 75.4 & 77.8 & 77.8 & Claude Code + Opus 5.5 \\
covid-\allowbreak{}hospitalization-\allowbreak{}forecast & 1 & rebuilt & 29.6 & 38.4 & 33.2 & 39.6 & Biomni + Gemini 3.8 Flash \\
cptac-\allowbreak{}pan-\allowbreak{}protein-\allowbreak{}imputation & 2 & rebuilt & 32.0 & 42.8 & 40.2 & 42.8 & Codex CLI + GPT-6 Astra \\
gutcrc-\allowbreak{}cross-\allowbreak{}cohort-\allowbreak{}detection & 1 & rebuilt & 56.1 & 55.7 & 54.8 & 59.3 & MLEvolve + DeepSeek V4.1 Flash \\
msk-\allowbreak{}met-\allowbreak{}tumor-\allowbreak{}origin & 1 & rebuilt & 56.9 & 59.8 & 59.9 & 60.4 & Gemini CLI + Gemini 3.8 Flash \\
survpath-\allowbreak{}multimodal & 1 & rebuilt & 56.4 & 60.4 & 59.3 & 60.4 & Claude Code + Opus 5 \\
target-\allowbreak{}aml-\allowbreak{}cytogenetics & 2 & rebuilt & 54.5 & 70.3 & 66.7 & 72.3 & Gemini CLI + Gemini 3.8 Flash \\
tcga-\allowbreak{}pancan-\allowbreak{}multiomics-\allowbreak{}survival & 2 & rebuilt & 59.1 & 61.9 & 60.6 & 61.9 & Codex CLI + GPT-6 Astra \\
\multicolumn{8}{@{}l}{\textbf{Sequence and genomics (8 tasks)}} \\
yeast-\allowbreak{}promoter-\allowbreak{}expression$^{\ast}$ & 1 & close & 82.1 & 76.0 & 48.0 & 80.4 & OpenCode + Opus 5.5 \\
deepstarr-\allowbreak{}genomic-\allowbreak{}sequence$^{\ast}$ & 1 & close & 60.7 & 55.7 & 54.8 & 59.7 & OpenCode + GPT-6 Astra \\
de-\allowbreak{}novo-\allowbreak{}peptide-\allowbreak{}sequencing-\allowbreak{}from-\allowbreak{}tandem-\allowbreak{}mass-\allowbreak{}spectrometry$^{\ast}$ & 1 & close & 73.4 & 74.5 & 67.4 & 74.5 & Codex CLI + GPT-6 Astra \\
proteingym-\allowbreak{}fitness$^{\ast}$ & 1 & close & 59.8 & 63.7 & 62.4 & 63.7 & Codex CLI + GPT-6 Astra \\
escape-\allowbreak{}amp-\allowbreak{}multilabel & 1 & rebuilt & 22.1 & 36.6 & 38.1 & 38.4 & OpenCode + Opus 5.5 \\
sirna-\allowbreak{}knockdown-\allowbreak{}efficacy & 1 & mixed & 53.1 & 57.3 & 56.5 & 59.2 & OpenCode + Opus 5.5 \\
aso-\allowbreak{}knockdown-\allowbreak{}efficacy & 1 & rebuilt & 26.8 & 36.2 & 33.4 & 36.4 & Claude Code + Opus 5 \\
cgbench-\allowbreak{}variant & 2 & rebuilt & 36.1 & 37.8 & 33.9 & 39.6 & Claude Code + Opus 5 \\
\multicolumn{8}{@{}l}{\textbf{Single-cell and spatial (8 tasks)}} \\
dlpfc-\allowbreak{}spatial-\allowbreak{}domain & 2 & rebuilt & 65.1 & 64.9 & 62.1 & 75.3 & MLEvolve + Gemini 3.8 Flash \\
gastrulation-\allowbreak{}kinetic-\allowbreak{}state & 2 & rebuilt & 69.4 & 81.5 & 74.0 & 87.1 & Biomni + Gemini 3.8 Flash \\
hest-\allowbreak{}coad-\allowbreak{}spatial-\allowbreak{}expression$^{\ast}$ & 1 & close & 19.0 & 21.2 & 22.1 & 25.2 & Claude Code + Opus 5 \\
lung-\allowbreak{}cite-\allowbreak{}cell-\allowbreak{}matching & 2 & rebuilt & 11.7 & 33.2 & 15.2 & 33.2 & Codex CLI + GPT-6 Astra \\
openproblems-\allowbreak{}rna-\allowbreak{}atac & 2 & rebuilt & 25.9 & 39.0 & 36.7 & 41.6 & OpenCode + GPT-6 Astra \\
openproblems-\allowbreak{}rna-\allowbreak{}protein & 1 & rebuilt & 36.1 & 38.7 & 38.4 & 38.7 & Codex CLI + GPT-6 Astra \\
seqfish-\allowbreak{}deconvolution & 1 & rebuilt & 59.0 & 66.1 & 65.7 & 66.1 & OpenCode + Opus 5.5 \\
tabula-\allowbreak{}sapiens-\allowbreak{}ood-\allowbreak{}annotation & 1 & rebuilt & 55.9 & 67.9 & 67.1 & 73.7 & MLEvolve + DeepSeek V4.1 Flash \\
\multicolumn{8}{@{}l}{\textbf{Structure (8 tasks)}} \\
cath-\allowbreak{}inverse-\allowbreak{}folding$^{\ast}$ & 1 & close & 50.0 & 51.3 & 50.6 & 51.3 & Codex CLI + GPT-6 Astra \\
psbench-\allowbreak{}multimer-\allowbreak{}mqa & 1 & rebuilt & 41.7 & 54.7 & 53.9 & 59.4 & Biomni + Opus 5.5 \\
decoydb-\allowbreak{}pose-\allowbreak{}classify & 2 & rebuilt & 34.8 & 54.2 & 47.1 & 54.2 & Codex CLI + GPT-6 Astra \\
skempi-\allowbreak{}ppi-\allowbreak{}ddg$^{\ast}$ & 1 & close & 37.1 & 36.6 & 41.1 & 41.1 & Claude Code + Opus 5.5 \\
asep-\allowbreak{}epitope & 1 & rebuilt & 16.7 & 19.5 & 18.6 & 21.4 & OpenCode + GPT-6 Astra \\
ppi-\allowbreak{}interface-\allowbreak{}residues & 2 & rebuilt & 80.4 & 74.9 & 74.9 & 75.2 & OpenCode + Opus 5.5 \\
cpsea-\allowbreak{}cyclic-\allowbreak{}peptide-\allowbreak{}affinity & 2 & rebuilt & 70.5 & 71.9 & 72.7 & 72.7 & Claude Code + Opus 5.5 \\
denoise-\allowbreak{}cryo & 2 & rebuilt & 72.1 & 81.3 & 57.4 & 81.3 & Codex CLI + GPT-6 Astra \\
\midrule \textbf{Mean (64 tasks)} & & & 48.6 & 55.1 & 53.2 & 57.0 & \\
\textbf{Mean, published-method baseline (9 tasks)} & & & 54.7 & 54.9 & 51.0 & 57.0 & \\
\end{longtable}
}

\clearpage
\phantomsection\label{LastPage}
\renewcommand{\refname}{Supplementary References}
\renewcommand{\bibname}{Supplementary References}

\end{document}